%% file: main.tex
\documentclass[10pt,journal,compsoc]{IEEEtran}

\ifCLASSOPTIONcompsoc
  \usepackage[nocompress]{cite}
\else
  \usepackage{cite}
\fi

\ifCLASSINFOpdf
\else
\fi

\usepackage[utf8]{inputenc}

\usepackage{threeparttablex} 
\usepackage{booktabs} 
\usepackage{adjustbox}
\usepackage{rotating}
\usepackage{multirow}
\usepackage{longtable}
\usepackage{makecell}
\usepackage{amsfonts}
\usepackage{siunitx}
\usepackage{multirow}
\usepackage{enumerate}
\usepackage{booktabs}
\usepackage{amssymb}
\usepackage{calc}
\usepackage{paralist}
\usepackage[caption=false]{subfig}
\usepackage{pifont}
\usepackage{tikz}
\usetikzlibrary{fit,positioning}
\usepackage{graphicx}
\usepackage[justification=justified]{caption}
\usepackage{diagbox}
\usepackage{color}
\usepackage{colortbl}
\usepackage{soul}
\usepackage{url}
\usepackage{mathtools}
\usepackage{algorithm}
\usepackage{algorithmic}
\usepackage[normalem]{ulem}
\usepackage{enumitem}
\usepackage{numprint}
\npthousandsep{,}
\npdecimalsign{.}
\usepackage{setspace}
\usepackage{colortbl}
\usepackage{acronym}
\usepackage{bm}
\usepackage{array}
\newcolumntype{P}[1]{>{\raggedright\arraybackslash}p{#1}}
\definecolor{tabcolor}{RGB}{210, 227, 248}

\usepackage{epstopdf}

\begin{document}
\title{Retrieving and Reading : A Comprehensive Survey on Open-domain Question Answering}

\author{Fengbin Zhu, Wenqiang Lei*, Chao Wang, Jianming Zheng, Soujanya Poria, Tat-Seng Chua,
\IEEEcompsocitemizethanks{
\IEEEcompsocthanksitem *Corresponding author: Wenqiang Lei
\IEEEcompsocthanksitem Fengbin Zhu, Wenqiang Lei and Tat-Seng Chua are with National University of Singapore (NUS) \protect 
E-mail: zhfengbin@gmail.com, wenqianglei@gmail.com, dcscts@nus.edu.sg
\IEEEcompsocthanksitem Fengbin Zhu and Chao Wang are with 6ESTATES PTE LTD, Singapore \protect
E-mail: wangchao@6estates.com
\IEEEcompsocthanksitem Jianming Zheng is with National University of Defense Technology, China \protect E-mail: zhengjianming12@nudt.edu.cn
\IEEEcompsocthanksitem Soujanya Poria is with Singapore University of Technology and Design (SUTD) \protect
E-mail: sporia@sutd.edu.sg}
}

\IEEEtitleabstractindextext{%
\begin{abstract}
Open-domain Question Answering (OpenQA) is an important task in Natural Language Processing (NLP), which aims to answer a question in the form of natural language based on large-scale unstructured documents.
Recently, there has been a surge in the amount of research literature on OpenQA, particularly on techniques that integrate with neural Machine Reading Comprehension (MRC).
While these research works have advanced performance to new heights on benchmark datasets, they have been rarely covered in existing surveys on QA systems. 
In this work, we review the latest research trends in OpenQA, with particular attention to systems that incorporate neural MRC techniques. 
Specifically, we begin with revisiting the origin and development of OpenQA systems.
We then introduce modern OpenQA architecture named ``Retriever-Reader'' and analyze the various systems that follow this architecture as well as the specific techniques adopted in each of the components.
We then discuss key challenges to developing OpenQA systems and offer an analysis of benchmarks that are commonly used.
We hope our work would enable researchers to be informed of the recent advancement and also the open challenges in OpenQA research, so as to stimulate further progress in this field.
\end{abstract}

}

\maketitle
%\IEEEdisplaynontitleabstractindextext
%\IEEEpeerreviewmaketitle

\IEEEraisesectionheading{\section{Introduction}\label{sec:intro}}
\input{01_Introduction}

\section{Development of OpenQA}\label{sec:advances}
\input{02_OpenQA}

\section{Modern OpenQA: Retrieving and Reading}\label{sec:openqa}
\input{03_Modern}

\section{Challenges and Benchmarks}\label{sec:challenge}
\input{04_Challenge}

\section{Conclusion}\label{sec:conclusion}
\input{05_Conclusion.tex}

\section{Acknowledgements}

This research is supported by the National Research Foundation, Singapore under its International Research Centres in Singapore Funding Initiative and A*STAR under its RIE 2020 Advanced Manufacturing and Engineering (AME) programmatic grant, Award No. - A19E2b0098, Project name - K-EMERGE: Knowledge Extraction, Modelling, and Explainable Reasoning for General Expertise. Any opinions, findings and conclusions or recommendations expressed in this material are those of the author(s) and do not reflect the views of National Research Foundation and A*STAR, Singapore.

\bibliographystyle{IEEEtran}
\bibliography{main.bib}

\begin{IEEEbiography}[{\includegraphics[width=1in,height=1.25in,clip,keepaspectratio]{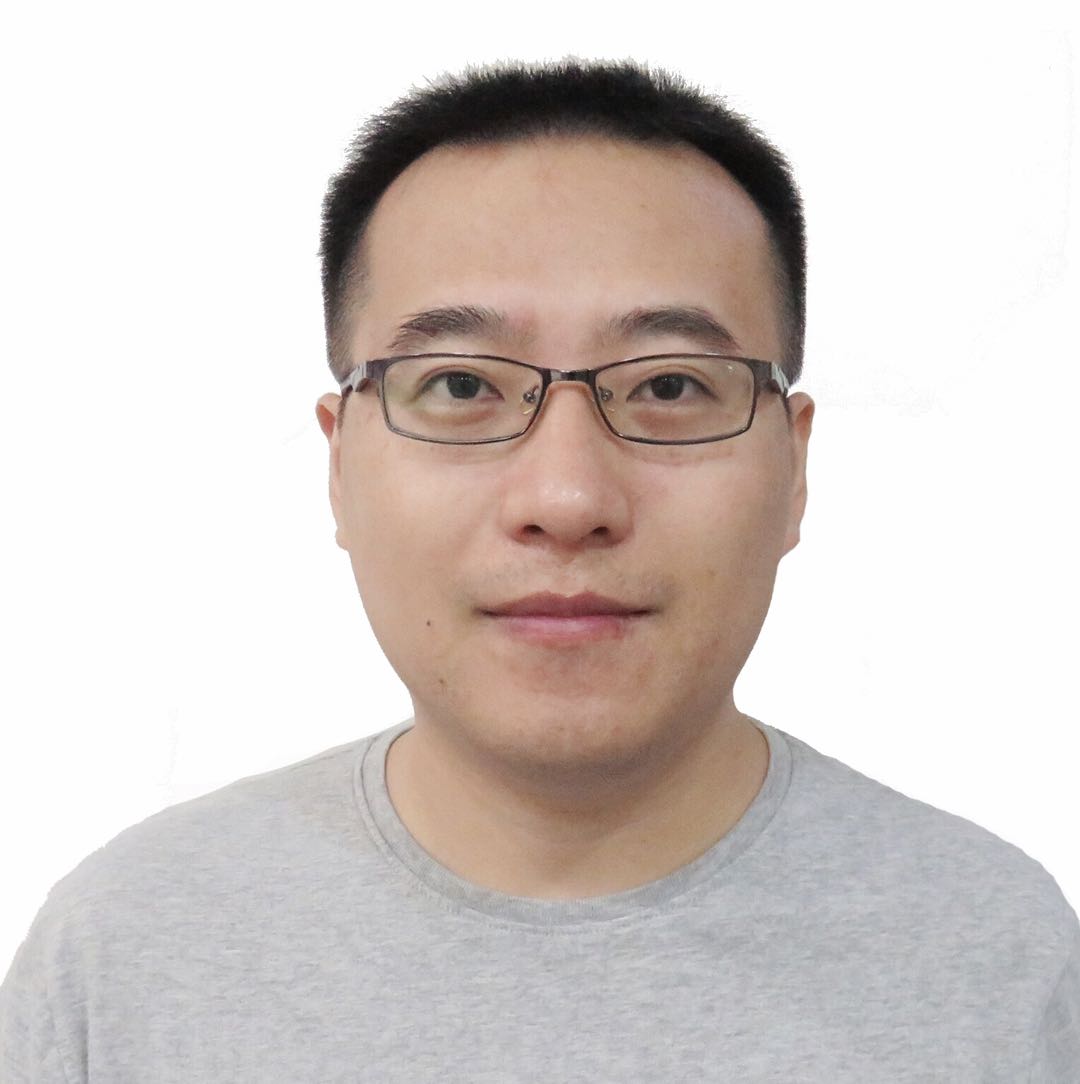}}]{Fengbin Zhu}  received his B.E. degree from Shandong University, China. He is currently pursuing his Ph.D degree at the School of Computing, National University of Singapore (NUS). 
His research interests include natural language processing, machine reading comprehension and conversational question answering.
\end{IEEEbiography}

\begin{IEEEbiography}[{\includegraphics[width=1in,height=1.25in,clip,keepaspectratio]{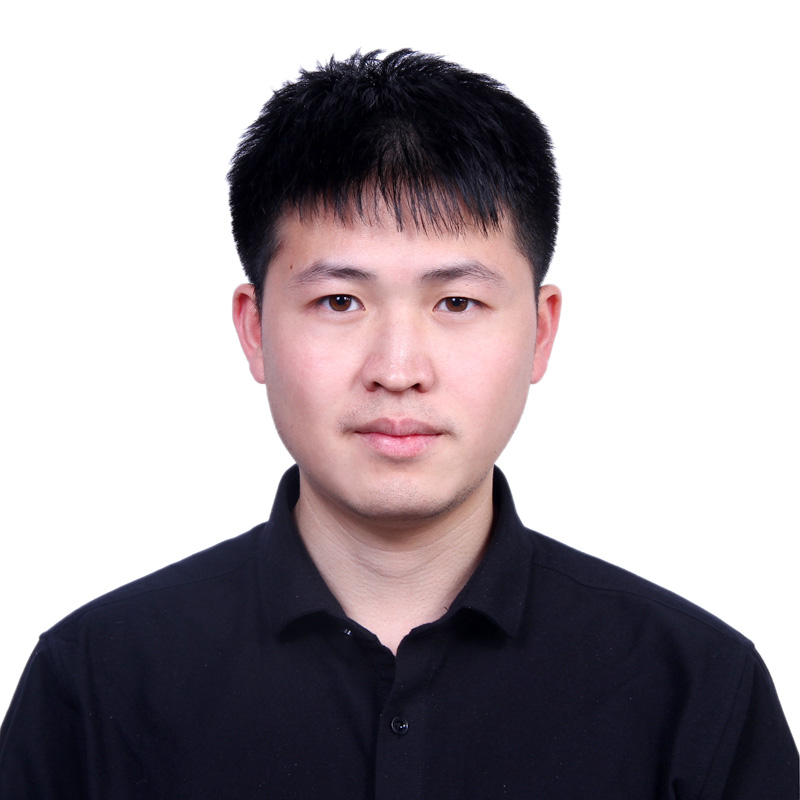}}]{Wenqiang Lei} is a Research Fellow with School of Computing, National University of Singapore (NUS). He received his Ph.D. in Computer Science from NUS in 2019. His research interests cover natural language processing and information retrieval, particularly on dialogue systems, conversational recommendations and question answering. He has published multiple papers at top conferences like ACL, IJCAI, AAAI, EMNLP and WSDM and the winner of ACM MM 2020 best paper award. He served as (senior) PC members on toptier conferences including ACL, EMNLP, SIGIR, AAAI, KDD and he is a reviewer for journals like TOIS, TKDE, and TASLP.
\end{IEEEbiography}

\begin{IEEEbiography}[{\includegraphics[width=1in,height=1.25in,clip,keepaspectratio]{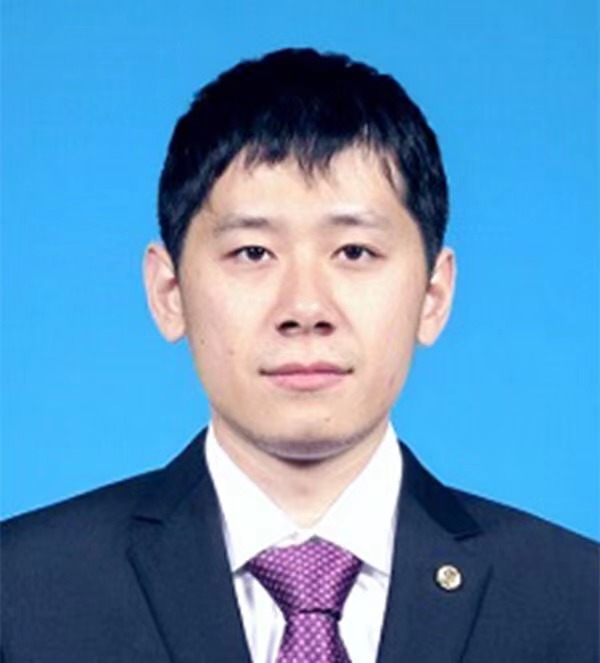}}]{Chao Wang}  holds a PhD in Computer Science from Tsinghua University, where he was advised by Dr. Shaoping Ma and Dr. Yiqun liu. His work has primarily focused on nature language processing, information retrieval, search engine user behavior analysis. His work has appeared in major journals and conferences such as SIGIR, CIKM, TOIS, and IRJ.
\end{IEEEbiography}

\begin{IEEEbiography}[{\includegraphics[width=1in,height=1.25in,clip,keepaspectratio]{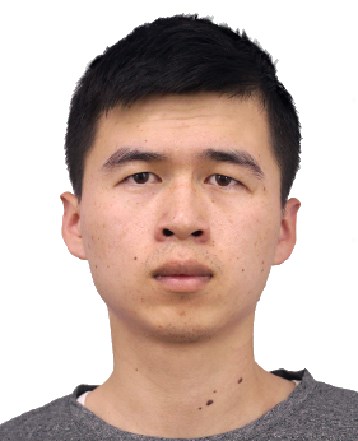}}]{Jianming Zheng}
is a PhD  candidate at the School of System Engineering, the National University of Defense Technology, China.
His research interests include  semantics representation, few-shot learning and its applications in information retrieval.
He received the BS and MS degrees from the National University of Defense Technology, China, in 2016 and 2018, respectively. 
He has several papers published in SIGIR, COLING, IPM, FITEE, Cognitive Computation, etc.
\end{IEEEbiography}

\begin{IEEEbiography}[{\includegraphics[width=1in,height=1.25in,clip,keepaspectratio]{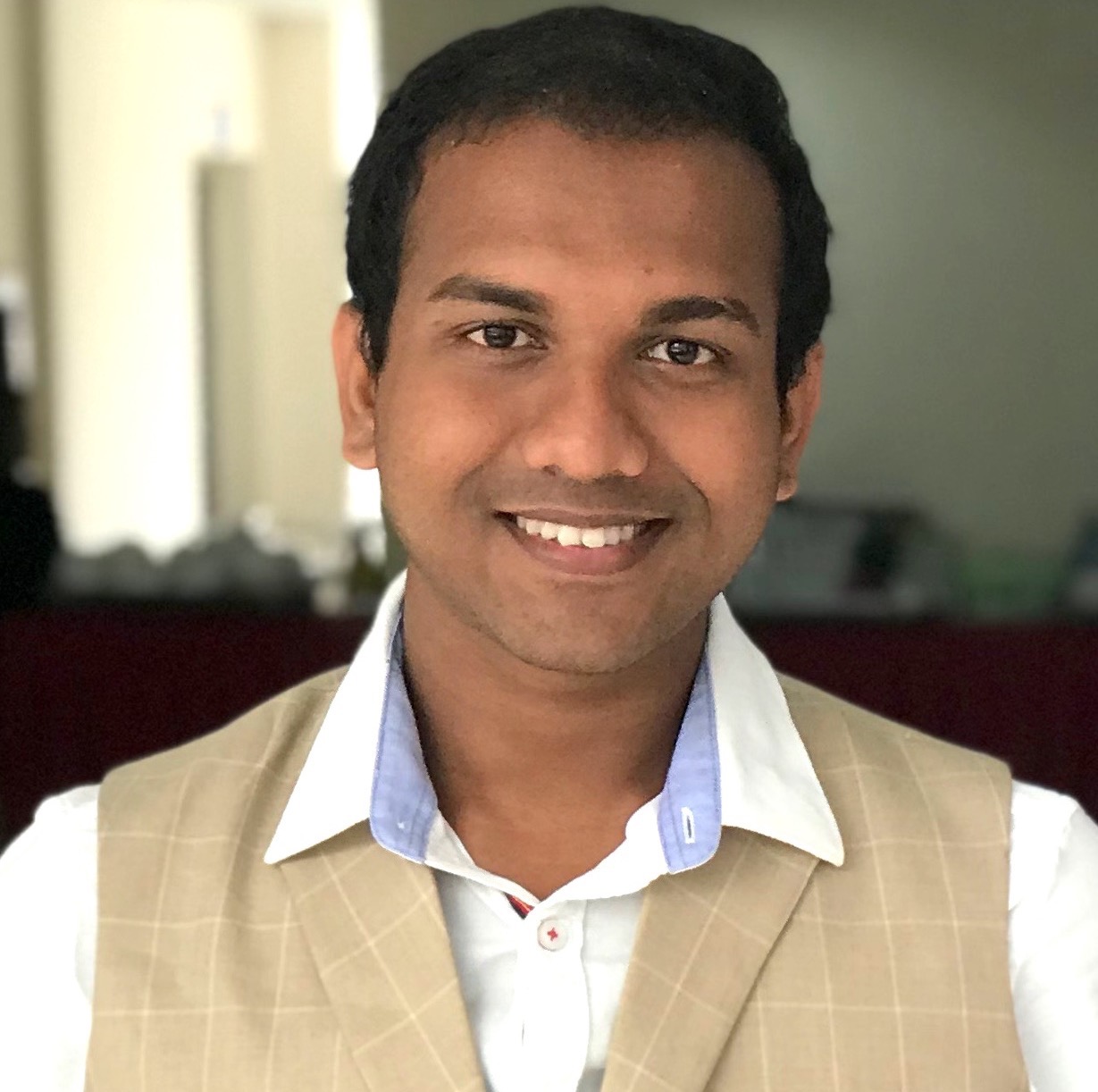}}]{Soujanya Poria} is an assistant professor of Information Systems Technology and Design, at the Singapore University of Technology and Design (SUTD), Singapore. He holds a Ph.D. degree in Computer Science from the University of Stirling, UK. He is a recipient of the prestigious early career research award called ``NTU Presidential Postdoctoral Fellowship" in 2018. Soujanya has co-authored more than 100 research papers, published in top-tier conferences and journals such as ACL, EMNLP, AAAI, NAACL, Neurocomputing, Computational Intelligence Magazine, etc. Soujanya has been an area chair at top conferences such as ACL, EMNLP, NAACL. Soujanya serves or has served on the editorial boards of the Cognitive Computation and Information Fusion.
\end{IEEEbiography}

\begin{IEEEbiography}[{\includegraphics[width=1in,height=1.25in,clip,keepaspectratio]{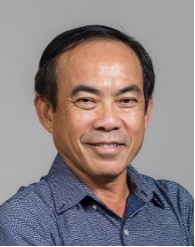}}]{Tat-Seng Chua}   is the KITHCT Chair Professor at the School of Computing, National University of Singapore (NUS). He is also the Distinguished Visiting Professor of Tsinghua University. Dr. Chua was the Founding Dean of the School of Computing from 1998-2000. His main research interests include heterogeneous data analytics, multimedia information retrieval, recommendation and conversation systems, and the emerging applications in E-commerce, wellness and Fintech. Dr. Chua is the co-Director of NExT, a joint research Center between NUS and Tsinghua, focusing on Extreme Search.

Dr. Chua is the recipient of the 2015 ACM SIGMM Achievements Award for the Outstanding Technical Contributions to Multimedia Computing, Communications and Applications. He is the Chair of steering committee of ACM ICMR (2015-19), and Multimedia Modeling (MMM) conference series. He was the General Co-Chair of ACM Multimedia 2005, ACM CIVR (now ACM ICMR) 2005, ACM SIGIR 2008, and ACM Web Science 2015. He serves in the editorial boards of three international journals. He holds a PhD from the University of Leeds, UK.
\end{IEEEbiography}

\end{document}

%% file: 01_Introduction.tex
% What is QA system
Question Answering (QA) aims to provide precise answers in response to the user's questions in natural language.
It is a long-standing task dating back to the 1960s\cite{Green1961Baseball}.
Compared with a search engine, the QA system aims to present the final answer to a question directly instead of returning a list of relevant snippets or hyperlinks, thus offering better user-friendliness and efficiency.
Nowadays many web search engines like Google and Bing have been evolving towards higher intelligence by incorporating QA techniques into their search functionalities~\cite{JOEL2011Google}.
Empowered with these techniques, search engines now have the ability to respond precisely to some types of questions such as 
\\\\
\emph{
--- Q: \textsl{``When was Barack Obama born?''}}
\\
\emph{
--- A: \textsl{``4 August 1961''}.
}
\\

\begin{figure*}[htb]
    \begin{center}
    \includegraphics[width=\textwidth]{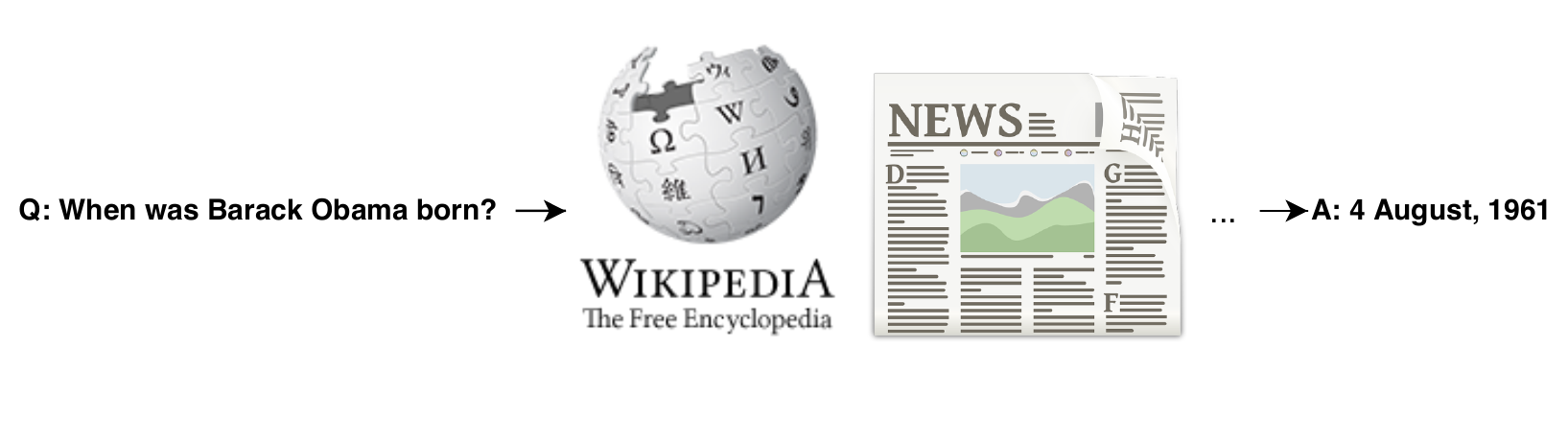}
    \end{center}
    \vspace{-1em}
    \caption{\label{fig:openqa_system} An illustration of OpenQA. Given a natural language question, the system infers the answer from a collection of unstructured text documents.}
\end{figure*}

The whole QA landscape can roughly be divided into two parts: textual QA and Knowledge Base (KB)-QA, according to the type of information source where answers are derived from. Textual QA mines answers from unstructured text documents while KB-QA from a predefined structured KB that is often manually constructed.  
Textual QA is generally more scalable than the latter, since most of the unstructured text resources it exploits to obtain answers are fairly common and easily accessible, such as Wikipedia~\cite{Chen2017Reading}, news articles~\cite{Voorhees99Trec8} and science books~\cite{Todor2018OpenBookQA}, etc.
Specifically, textual QA is studied under two task settings based on the availability of contextual information, i.e. Machine Reading Comprehension (MRC) and Open-domain QA (OpenQA).
MRC, which originally took inspiration from language proficiency exams, aims to enable machines to read and comprehend specified context passage(s) for answering a given question.
In comparison, OpenQA tries to answer a given question without any specified context.
It usually requires the system to first search for the relevant documents as the context w.r.t. a given question from either a local document repository or the World Wide Web (WWW), and then generate the answer, as illustrated in Fig.~\ref{fig:openqa_system}.
OpenQA therefore enjoys a wider scope of application and is more in line with real-world QA behavior of human beings while MRC can be considered as a step to OpenQA~\cite{Harabagiu2003Open}.
In fact, building an OpenQA system that is capable of answering any input questions is deemed as the ultimate goal of QA research.

% Traditional OpenQA 
In literature, OpenQA has been studied closely with research in Natural Language Processing (NLP), Information Retrieval (IR), and Information Extraction (IE)~\cite{Burger2001Issues, Kolomiyets2011SQASurvey, Allam2012QASurvey, Mishra2016QASurvey}.
Traditional OpenQA systems mostly follow a pipeline consisting of three stages, i.e. \emph{Question Analysis}, \emph{Document Retrieval} and \emph{Answer Extraction}~\cite{Harabagiu2003Open,Pasca2003Open,  Allam2012QASurvey}.
Given an input question in natural language, \emph{Question Analysis} aims to reformulate the question to generate search queries for facilitating subsequent \emph{Document Retrieval} and classify the question to obtain its expected answer type(s) that would guide \emph{Answer Extraction}. 
In the \emph{Document Retrieval} stage, the system searches for question-relevant documents or passages with the generated search queries, usually using existing IR techniques like TF-IDF and BM25, or specific techniques developed for Web search engines like Google.com and Bing.com.
After that, in the \emph{Answer Extraction} stage, the final answer is extracted from related documents received from the previous stage.

% Modern OpenQA
Deep learning techniques, which have driven remarkable progress in many fields, have also been successfully applied to  almost every stage of OpenQA systems \cite{Huang2020Recent}.
For example, \cite{Tao2018Novel} and~\cite{Xia2018Novel} develop the question classifier using a CNN-base model and an LSTM-based model respectively. 
In \cite{Nishida2018Retrieve,Karpukhin2020Dense,Khattab2020Relevance}, they propose neural retrieval models to search for relevant documents in a latent space.
In recent years, with the emergence of some large-scale QA datasets\cite{Hermann2015Teaching,Rajpurkar2016SQuAD,Nguyen2016MS,Lai2017RACE,Rajpurkar2018Know,li2020molweni}, neural MRC techniques have been greatly advanced~\cite{Hermann2015Teaching, Seo2017Bidirectional,Wang2017Gated,Yu2018QANET,Devlin2018Bert}.
By adopting the popular neural MRC methods to extract the answer to a given question from the relevant document(s), traditional OpenQA systems have been revolutionized~\cite{Chen2017Reading, Wang2018R3RR, Das2019Multistep, Guu2020Realm} and evolved to a modern  \emph{``Retriever-Reader''} architecture.
\emph{Retriever} is responsible for retrieving relevant documents w.r.t. a given question, which can be regarded as an IR system, while \emph{Reader} aims at inferring the final answer from the received documents, which is usually a neural MRC model.
A handful of works~\cite{Chen2017Reading, Ding2019Cognitive, Nie2019Revealing} even re-name OpenQA as Machine Reading Comprehension at Scale (MRS).
Following this architecture, extensive research has been made along various directions, such as re-ranking the retrieved documents before feeding them into a neural MRC model~\cite{Wang2018R3RR, Lin2018Denoising, Lee2018Ranking}, retrieving the relevant documents iteratively given a question~\cite{Das2019Multistep, Feldman2019Multihop, Qi2019Answering}, and training the entire OpenQA system in an end-to-end manner~\cite{Nishida2018Retrieve, Lee2019Latent, Guu2020Realm, Lewis2020Retrieval}, etc.

Based on above observations and insights, we believe it is time to provide a comprehensive literature review on OpenQA systems, with particular attention to techniques that incorporate neural MRC models.
Our review is expected to acknowledge the advancement that has been made thus far and summarize the current challenges to stimulate further progress in this field.
In the rest of this survey, we will present the following contents. 
In Section~\ref{developmentofopenqa}, we review the development of OpenQA systems, including the origin, traditional architecture, and recent progress in using deep neural networks. 
In Section~\ref{modernopenqa}, we summarize and elaborate a \emph{``Retriever-Reader''} architecture for OpenQA followed by detailed analysis on the various techniques adopted.
In Section~\ref{challengesandbenchmarks}, we first discuss some salient challenges towards OpenQA, identifying the research gaps and hoping to enhance further research in this field, and subsequently provide a summary and analysis of QA benchmarks that are applicable to either MRC or OpenQA.
Finally, we draw our conclusions based on the presented contents above in Section~\ref{conclusion}.

%% file: 02_OpenQA.tex
\label{developmentofopenqa}

In this section, we first briefly introduce the origin of Open-domain Question Answering (OpenQA) and then review the traditional and deep learning approaches to OpenQA sequentially to describe its remarkable advancement in the past two decades.

\subsection{Origin of OpenQA}
Pioneering research on Question Answering (QA) system was conducted within the scope of Information Retrieval (IR), with the focus on restricted domain or closed-domain settings. 
The earliest QA system, as is widely acknowledged, is the Baseball~\cite{Green1961Baseball}, which was designed in 1961 to answer questions about American baseball games, such as game time, location and team name.
Within this system, all the relevant information is stored in a well-defined dictionary, and user questions are translated into query statements using linguistic methods to extract final answers from the dictionary. 
In 1973, another famous QA system LUNAR~\cite{Woods1973LUNAR} was developed as a powerful tool to assist research work of lunar geologists, where chemical analysis data about lunar rocks and soil obtained from Apollo moon missions are stored in one data base file provided by NASA MSC for each scientist to conveniently view and analyze.
In 1993, MURAX~\cite{Kupiec1993MURAX} was designed to answer simple academic questions based on an English academic encyclopedia which mainly employs linguistic analysis and syntactic pattern matching techniques.

In 1999, OpenQA was first defined as extracting top 5 probable snippets containing the correct answer from a collection of news articles in the QA track launched by the Text REtrieval Conference (TREC)~\cite{Voorhees99Trec8}.
Compared to the previous research on QA, in the open-domain setting, a large number of unstructured documents is used as the information source from which the correct answer to a given question would be extracted.
In the later years, a series of TREC QA Tracks have remarkably advanced the research progress on OpenQA~\cite{Voorhees2001Trec,Voorhees2003Trec11,Voorhees2003Trec12}.
It is worth noting that systems are required to return exact short answers to given questions starting from TREC-11 held in 2002~\cite{Voorhees2003Trec11}.

% Web 
The TREC campaign provides a local collection of documents as the information source for generating answers, but the popularity of World Wide Web (WWW), especially the increasing maturity of search engines, has inspired researchers to build Web-based OpenQA systems~\cite{Kupiec1993MURAX, Kwok2001Scaling, Brill2002AskMSR, Zheng2002AnswerBusQA} obtaining answers from online resources like Google.com and Ask.com, using IR techniques.   
Web search engines are able to consistently and efficiently collect massive web pages, therefore capable of providing much more information to help find answers in response to user questions. 
In 2001, a QA system called MULDER~\cite{Kwok2001Scaling} was designed to automatically answer open-domain factoid questions with a search engine (e.g., Google.com). 
It first translates users' questions to multiple search queries with several natural-language parsers and submits them to the search engine to search for relevant documents, and then employs an answer extraction component to extract the answer from the returned results.
Following this pipeline, a well-known QA system AskMSR~\cite{Brill2002AskMSR} was developed, which mainly depends on data redundancy rather than sophisticated linguistic analysis of either questions or candidate answers.
It first translates the user's question into queries relying on a set of predefined rewriting rules to gather relevant documents from search engines and then adopts a series of n-gram based algorithms to mine, filter and select the best answer.
For such OpenQA systems, the search engines are able to provide access to an ocean of information, significantly enlarging the possibility of finding precise answers to user questions.
Nevertheless, such an ample information source also brings considerable noisy content that challenges the QA system to filter out.

\subsection{Traditional Architecture of OpenQA}
\begin{figure*}[t]
  \includegraphics[width=0.9\textwidth]{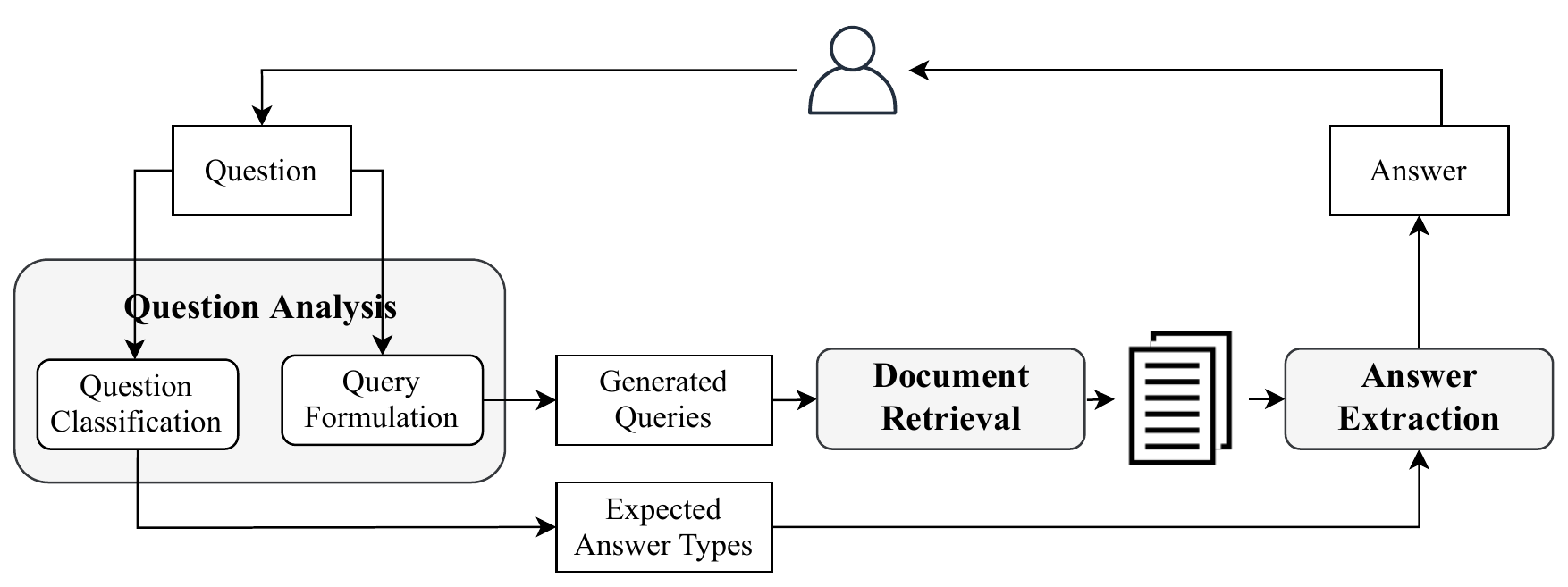}
  \centering
  \caption{An illustration of traditional architecture of OpenQA system}
  \label{fig:typical}
\end{figure*}

The traditional architecture of OpenQA systems is illustrated in Fig. \ref{fig:typical}, which mainly comprises three stages: \emph{Question Analysis}, \emph{Document Retrieval}, and \emph{Answer Extraction}~\cite{Harabagiu2003Open,Pasca2003Open}.
Given a natural language question, \emph{Question Analysis} aims to understand the question first so as to facilitate document retrieval and answer extraction in the following stages.
Performance of this stage is found to have a noticeable influence upon that of the following stages, and hence important to the final output of the system~\cite{Moldovan2002Performance}.
Then, \emph{Document Retrieval} stage searches for question-relevant documents based on a self-built IR system~\cite{Voorhees99Trec8} or Web search engine~\cite{Kwok2001Scaling, Brill2002AskMSR} using the search queries generated by \emph{Question Analysis}.
Finally, \emph{Answer Extraction} is responsible for extracting final answers to user questions from the relevant documents received in the preceding step.
In the following, we will analyze each stage one by one.

\subsubsection{Question Analysis}
The goals of \emph{Question Analysis} stage are two-fold.
On one hand, it aims to facilitate the retrieval of question-relevant documents, for which a Query Formulation module is often adopted to generate search queries. 
On the other hand, it is expected to enhance the performance of \emph{Answer Extraction} stage by employing a Question Classification module to predict the type of the given question, which leads to a set of expected answer types.
A simple illustration of this stage is given in the leftmost grey box of Fig. \ref{fig:typical}.

In Query Formulation, 
linguistic techniques such as POS tagging~\cite{Kupiec1993MURAX, Kwok2001Scaling}, stemming~\cite{Kupiec1993MURAX}, parsing~\cite{Kwok2001Scaling} and stop word removal~\cite{Brill2002AskMSR, Sun2005Using} are usually utilized to extract keywords for retrieving.
However, the terms used in questions are often not the same as those appearing in the documents that contain the correct answers. 
This problem is called ``term mismatch'' and is a long-standing and critical issue in IR.
To address this problem, query expansion~\cite{Xu1996QELG, Carpineto2012SurveyAQE} and paraphrasing techniques~\cite{Quirk2004Monolingual, Bannard2005Paraphrasing,Zhao2008Combining,Wubben2010Paraphrase}
are often employed to produce additional search words or phrases so as to retrieve more relevant documents.

Question Classification, the other module that is often adopted for \emph{Question Analysis} stage, aims to identify the type of the given question based on a set of question types (e.g., where, when, who, what) or a taxonomy~\cite{Li2002Learning, Suzuki2003Question} manually defined by linguistic experts.
After obtaining the type of the question, expected answer types can be easily determined using rule-based mapping methods~\cite{Allam2012QASurvey}.
For example, given a question \textsl{``When was Barack Obama born?''},  the answer type would be inferred as \textsl{``Date''} when knowing the question type is \textsl{``When''}. 
Identifying the question type can provide constraint upon answer extraction and significantly reduce the difficulty of finding correct answers.
Question Classification has attracted much interest in literature~\cite{Kwok2001Scaling,Li2002Learning, Zhang2003Question, Ferrucci2010Watson, Madabushi2016High}.
For instance, \cite{Madabushi2016High} proposed to extract relevant words from a given question and then classify the question based on rules associating these words to concepts;
\cite{Zhang2003Question} trained a list of question classifiers using various machine learning techniques such as Support Vector Machines (SVM), Nearest Neighbors and Decision Trees on top of the hierarchical taxonomy proposed by~\cite{Li2002Learning}.

\subsubsection{Document Retrieval}
\label{sec:document-processing}
This stage is aimed at obtaining a small number of relevant documents that probably contain the correct answer to a given question from a collection of unstructured documents, which usually relies on an IR engine.
It can significantly reduce the search space for arriving at the final answer.

%document retrieval
In the past decades, various retrieval models have been developed for \emph{Document Retrieval}, among which some popular ones are the Boolean model, Vector Space Models, Probabilistic Models, Language Models \cite{Manning2008IR}, etc., which are briefly revisited as follows. 
\begin{itemize}
\item \textit{Boolean Model}: The Boolean Model is one of the simplest retrieval models.
The question is transformed into the form of a Boolean expression of terms, which are combined with the operators like "AND", "OR" and "NOT" to exactly match with the documents, with each document  viewed as a set of words.
\item \textit{Vector Space Model}: The Vector Space Models represent the question and each document as word vectors in a \textsl{d}-dimensional word space, where \textsl{d} is the number of words in the vocabulary.
When searching for relevant documents to a given question, the relevance score of each document is computed by computing the similarity (e.g., the cosine similarity) or distance (e.g., the euclidean distance) between its vector and the question vector.
Compared to the Boolean model, this approach returns documents to the question even if the constraints posed by the question are only partially met, with precision sacrificed.
\item \textit{Probabilistic Model}: The Probabilistic Models provide a way of integrating probabilistic relationships between words into a model.
Okapi BM25~\cite{Robertson2009probabilistic} is a probabilistic model sensitive to term frequency and document length, which is one of the most empirically successful retrieval models and widely used in current search engines.
\item \textit{Language Model}: The Language Models~\cite{Kluwer2003Language} are also very popular, among which the Query Likelihood Model~\cite{Manning2008IR} is the most widely adopted.
It builds a probabilistic language model $LM_d$ for each document $d$ and ranks documents according to the probability $P(q \mid LM_d)$ of the language model generating the given question $q$.
\end{itemize}

%document post-processing
In practice, the documents received often contain irrelevant ones, or the number of documents is so large that the capacity of the \emph{Answer Extraction} model is overwhelmed.
To address the above issues, post-processing on the retrieved documents is very demanded.
Widely used approaches on processing retrieved documents include document filtering, document re-ranking and document selection~\cite{Allam2012QASurvey},  etc.
Document filtering is used to identify and remove the noise w.r.t. a given question; document re-ranking is developed to further sort the documents according to a plausibility degree of containing the correct answer in the descending order; document selection is to choose the top relevant documents.
After post-processing, only the most relevant documents would be remained and fed to the next stage to extract the final answer.

\subsubsection{Answer Extraction}
The ultimate goal of an OpenQA system is to successfully answer given questions, and \emph{Answer Extraction} stage is responsible for returning a user the most precise answer to a question.
The performance of this stage is decided by the complexity of the question, the expected answer types from \emph{Question Analysis} stage, the retrieved documents from \emph{Document Retrieval} stage as well as the extraction method adopted, etc.
With so many influential factors, researchers need to take a lot of care and place special importance on this stage.

In traditional OpenQA systems, factoid questions and list questions~\cite{Voorhees2004Trec} have been widely studied for a long time.
Factoid questions (e.g., When, Where, Who...) to which the answers are usually a single text span in in the documents, such as such as an entity name, a word token or a noun phrase. While list questions whose answers are a set of factoids that appeared in the same document or aggregated from different documents.
The answer type received from the stage of \emph{Question Analysis} plays a crucial role, especially for the given question whose answers are named entities.
Thus, early systems heavily rely on the Named Entity Recognition (NER) technique~\cite{Molla2006Named,Zheng2002AnswerBusQA,Kupiec1993MURAX} since comparing the recognised entities and the answer type may easily yield the final answer.
In \cite{Wang2006ASO}, the answer extraction is described as a unified process, first uncovering latent or hidden information from the question and the answer respectively, and 
then using some matching methods to detect answers, such as surface text pattern matching~\cite{Soubbotin2001Patterns,Ravichandran2002Learning}, word or phrase matching~\cite{Kwok2001Scaling}, and syntactic structure matching~\cite{Shen2005Exploring,Kupiec1993MURAX,Sun2005Using}.

In practice, sometimes the extracted answer needs to be validated when it is not confident enough before presenting to the end-users.
Moreover, in some cases multiple answer candidates may be produced to a question and we have to select one among them.
Answer validation is applied to solve such issues.
One widely applied validation method is to adopt an extra information source like a Web search engine to validate the confidence of each candidate answer. 
The principle is that the system should return a sufficiently large number of documents which contain both question and answer terms. 
The larger the number of such returned documents is, the more likely it will be the correct answer. 
This principle has been investigated and demonstrated fairly effective, though simple~\cite{Allam2012QASurvey}.

\subsection{Application of Deep Neural Networks in OpenQA}

In the recent decade, deep learning techniques have also been successfully applied to OpenQA.
In particular, deep learning has been used in almost every stage in an OpenQA system, and moreover, it enables OpenQA systems to be end-to-end trainable.
For \emph{Question Analysis}, some works develop neural classifiers to determine the question types.
For example, \cite{Tao2018Novel} and~\cite{Xia2018Novel} respectively adopt a CNN-based and an LSTM-based model to classify the given questions, both achieving competitive results.
For \emph{Document Retrieval}, dense representation based methods~\cite{Das2019Multistep, Feldman2019Multihop, Guu2020Realm,Karpukhin2020Dense} have been proposed to address  ``term-mismatch'', which is a long-standing problem that harms retrieval performance.
Unlike the traditional methods such as TF-IDF and BM25 that use sparse representations, deep retrieval methods learn to encode questions and documents into a latent vector space where text semantics beyond term match can be measured.
For example, \cite{Das2019Multistep} and \cite{Feldman2019Multihop} train their own encoders to encode each document and question independently into dense vectors, and the similarity score between them is computed using the inner product of their vectors.
The Sublinear Maximum Inner Product Search (MIPS) algorithm~\cite{Ram2012Maximum,Shrivastava2014ALSH,Shen2015Learning} is used to improve the retrieval efficiency given a question, especially when the document repository is large-scale.
For \emph{Answer Extraction}, as a decisive stage for OpenQA systems to arrive at the final answer, neural models can also be applied.
Extracting answers from some relevant documents to a given question essentially makes the task of Machine Reading Comprehension (MRC).
In the past few years, with the emergence of some large-scale datasets such as CNN/Daily Mail~\cite{Hermann2015Teaching}, MS MARCO~\cite{Nguyen2016MS}, RACE~\cite{Lai2017RACE} and SQuAD 2.0~\cite{Rajpurkar2018Know}, 
research on neural MRC has achieved remarkable progress~\cite{Seo2017Bidirectional,Wang2017Gated,Yu2018QANET,Devlin2018Bert}. 
For example, BiDAF~\cite{Seo2017Bidirectional} represents the given document at different levels of granularity via a multi-stage hierarchical structure consisting of a character embedding layer, a word embedding layer, and a contextual embedding layer, and leverages a bidirectional attention flow mechanism to obtain a question-aware document representation without early summarization.
QANet~\cite{Yu2018QANET} adopts CNN and the self-attention mechanism~\cite{Vaswani2017Attention} to model the local interactions and global interactions respectively, which performs significantly faster than usual recurrent models.

Furthermore, applying deep learning enables the OpenQA systems to be end-to-end trainable~\cite{Nishida2018Retrieve,Lee2019Latent,Guu2020Realm}.
For example, \cite{Lee2019Latent} argue it is sub-optimal to incorporate a standalone IR system in an OpenQA system, and they develop an ORQA system that treats the document retrieval from the information source as a latent variable and trains the whole system only from question-answer string pairs based on BERT~\cite{Devlin2018Bert}.
REALM~\cite{Guu2020Realm} is a pre-trained language model that contains a knowledge retriever and a knowledge augmented encoder.
Both its retriever and encoder are differentiable neural networks, which are able to compute the gradient w.r.t. the model parameters to be back propagated  all the way throughout the network.
Similar to other pre-training language models, it also has two stages, i.e., pre-training and fine-tuning.
In the pre-training stage, the model is trained in an unsupervised manner, using masked language modeling as the learning signal while the parameters are fine-tuned using supervised examples in the fine-tuning stage.

In early OpenQA systems, the success of answering a question is highly dependent on the performance of \emph{Question Analysis}, particularly Question Classification, that provides expected answer types~\cite{Moldovan2002Performance}.
However, either the types or the taxonomies of questions are hand-crafted by linguists, which are non-optimal since it is impossible to cover all question types in reality, especially those complicated ones.
Furthermore, the classification errors would easily result in the failure of answer extraction, thus severely hurting the overall performance of the system.
According to the experiments in \cite{Moldovan2002Performance}, about 36.4\% of errors in early OpenQA systems are caused by miss-classification of question types.
Neural models are able to automatically transform questions from natural language to representations that are more recognisable to machines.
Moreover, neural MRC models provide an unprecedented powerful solution to \emph{Answer Extraction} in OpenQA, largely offsetting the necessity of applying the traditional linguistic analytic techniques to questions and bringing revolutions to OpenQA systems~\cite{Chen2017Reading, Wang2018R3RR, Das2019Multistep, Lee2019Latent}.
The very first work to incorporate neural MRC models into the OpenQA system is DrQA proposed by~\cite{Chen2017Reading}, evolving to a ``Retriever-Reader'' architecture.
It combines TF-IDF based IR technique and a neural MRC model to answer open-domain factoid questions over Wikipedia and achieves impressive performance.
After~\cite{Chen2017Reading}, lots of works have been released~\cite{Wang2018R3RR, Lin2018Denoising, Lee2018Ranking,Lee2019Latent, Seo2019Real,Dehghani2019Learning,Dhingra2020Differentiable, Guu2020Realm}.
Nowadays, to build OpenQA systems following the ``Retriever-Reader'' architecture has been widely acknowledged as the most efficient and promising way, which is also the main focus of this paper.

%% file: 03_Modern.tex
\label{modernopenqa}

\begin{figure*}[!htb]
  \includegraphics[width=\textwidth]{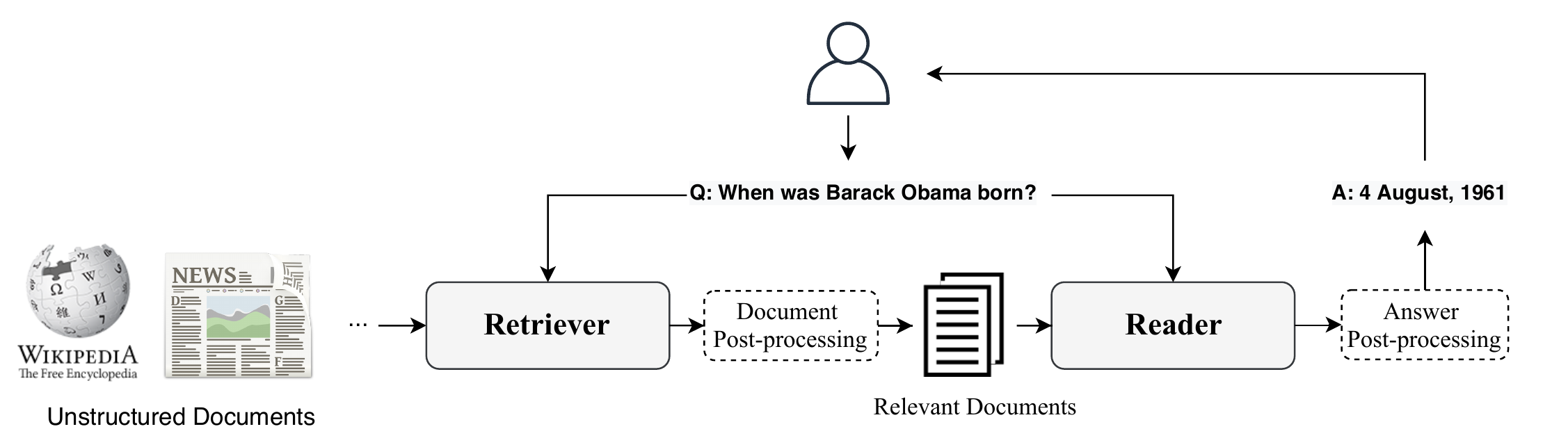}
  \caption{An illustration of ``Retriever-Reader'' architecture of OpenQA system. 
  The modules marked with dash lines are auxiliary.
  } 
  \label{fig:modern}
\end{figure*}

\begin{figure*}[!ht]
  \includegraphics[width=0.9\textwidth]{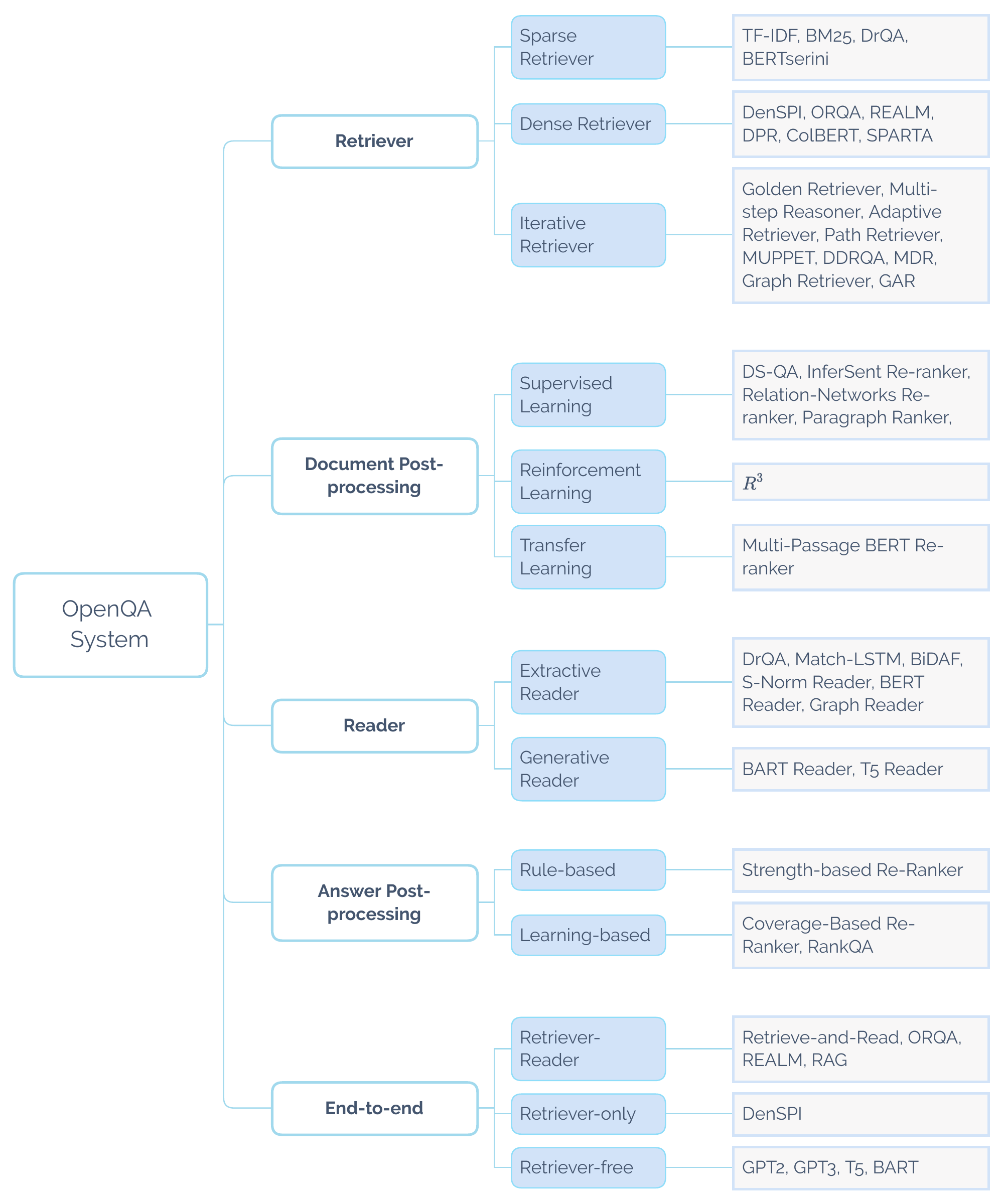}
  \caption{A taxonomy of \emph{``Retriever-Reader''} OpenQA system.
 } 
  \label{fig:taxonomy}
\end{figure*}

In this section, we introduce the \emph{``Retriever-Reader''} architecture of the OpenQA system, as illustrated in Fig. \ref{fig:modern}.
\emph{Retriever} is aimed at retrieving relevant documents w.r.t. a given question, which can be regarded as an IR system, while \emph{Reader} aims at inferring the final answer from the received documents, which is usually a neural MRC model.
They are two major components of a modern OpenQA system.
In addition, some other auxiliary modules, which are marked in dash lines in Fig.~\ref{fig:modern}, can also be incorporated into an OpenQA system, including \emph{Document Post-processing} that filters and re-ranks retrieved documents in a fine-grained manner to select the most relevant ones, and \emph{Answer Post-processing} that is to determine the final answer among multiple answer candidates.
The systems following this architecture can be classified into two groups, i.e. \emph{pipeline systems} and \emph{end-to-end systems}.
In the following, we will introduce each component with the respective approaches in the pipeline systems, then followed by the end-to-end trainable ones.
In Fig.~\ref{fig:taxonomy} we provide a taxonomy of the modern OpenQA system to make our descriptions better understandable.

\subsection{Retriever}
\emph{Retriever} is usually regarded as an IR system, with the goal of retrieving related documents or passages that probably contain the correct answer given a natural language question as well as ranking them in a descending order according to their relevancy.
Broadly, current approaches to \emph{Retriever} can be classified into three categories, i.e. \emph{Sparse Retriever}, \emph{Dense Retriever}, and \emph{Iterative Retriever}, which will be detailed in the following. 

\subsubsection{Sparse Retriever} 
It refers to the systems that search for the relevant documents by adopting classical IR methods as introduced in Section~\ref{sec:document-processing}, such as TF-IDF~\cite{Chen2017Reading,Kratzwald2019RankQA,Kratzwald2018Adaptive, Lee2018Ranking} and BM25~\cite{Yang2019End2end,Wang2019Multipassage}.
DrQA~\cite{Chen2017Reading} is the very first approach to modern OpenQA systems and developed by combining classical IR techniques and neural MRC models to answer open-domain factoid questions. 
Particularly, the retriever in DrQA adopts bi-gram hashing~\cite{Weinberger2009Feature} and TF-IDF matching to search over Wikipedia, given a natural language question.
BERTserini~\cite{Yang2019End2end} employs Anserini~\cite{Yang2017Anserini} as its retriever, which is an open-source IR toolkit based on Lucene.
In~\cite{Yang2019End2end}, different granularities of text including document-level, paragraph-level and sentence-level are investigated experimentally, and the results show paragraph-level index achieves the best performance.
Traditional retrieval methods such as TF-IDF and BM25 use sparse representations to measure term match.
However, the terms used in user questions are often not the same as those appearing in the documents. 
Various methods based on dense representations~\cite{Das2019Multistep, Feldman2019Multihop, Guu2020Realm,Karpukhin2020Dense} have been developed in recent years, which learn to encode questions and documents into a latent vector space where text semantics beyond term match can be measured.

\subsubsection{Dense Retriever} 
\begin{figure*}[ht]
  \includegraphics[width=0.9\textwidth]{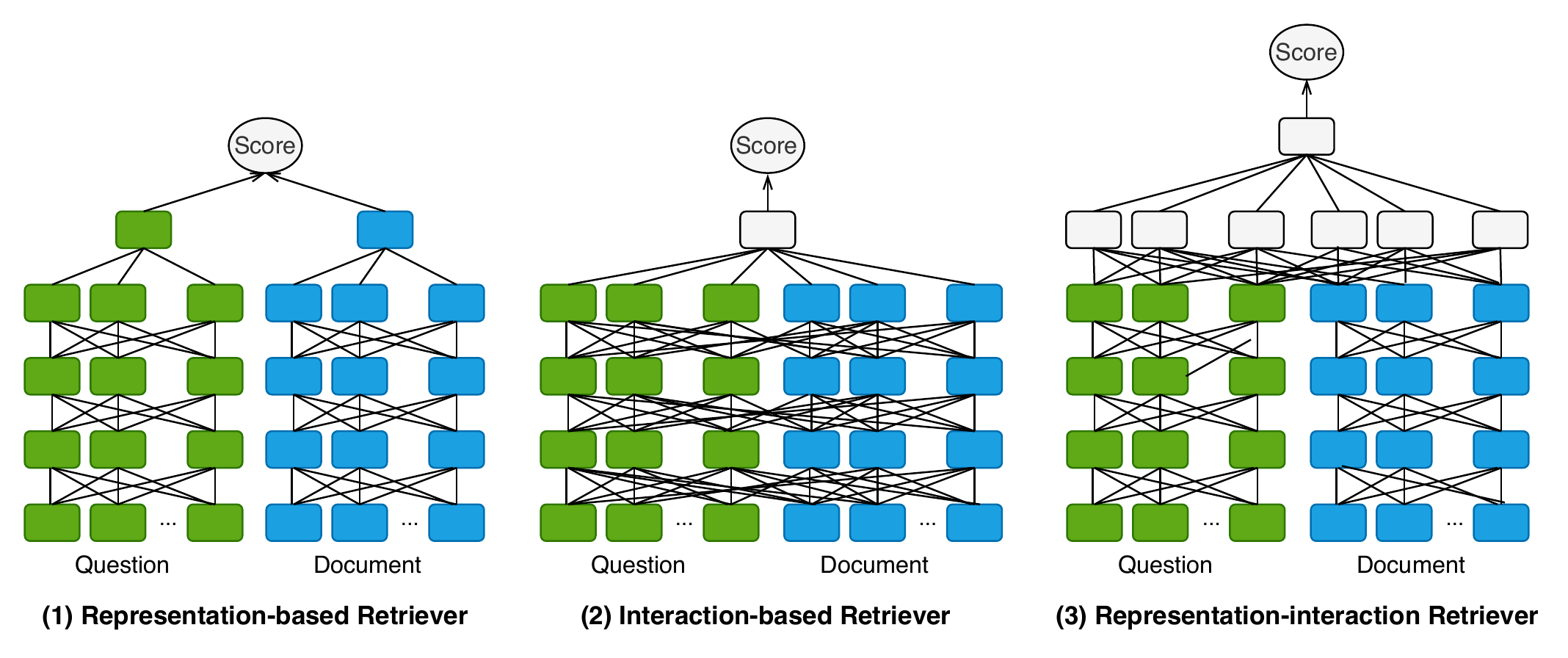}
  \centering
  \caption{Three types of dense retrievers.} 
  \label{fig:retriever}
\end{figure*}

Along with the success of deep learning that offers remarkable semantic representation, various deep retrieval models have been developed in the past few years, greatly enhancing retrieval effectiveness and thus lifting final QA performance.
According to the different ways of encoding the question and document as well as of scoring their similarity, dense retrievers in existing OpenQA systems can be roughly divided into three types: \emph{Representation-based Retriever}~\cite{Seo2019Real,Lee2019Latent,Guu2020Realm,Karpukhin2020Dense}, \emph{Interaction-based Retriever}~\cite{Nishida2018Retrieve,Nie2019Revealing}, and \emph{Representation-interaction Retriever}~\cite{Khattab2020Relevance,Zhao2020SPARTA,zhang2020dcbert}, as illustrated in Fig. \ref{fig:retriever}.

\textbf{Representation-based Retriever:} 
Representation-based Retriever, also called Dual-encoder or Two-tower retriever, employs two independent encoders like BERT~\cite{Devlin2018Bert} to encode the question and the document respectively, and estimates their relevance by computing a single similarity score between two representations.
For example, ORQA~\cite{Lee2019Latent} adopts two independent BERT-based encoders to encode a question and a document respectively and the relevance score between them is computed by the inner product of their vectors.
In order to obtain a sufficiently powerful retriever, they pre-train the retriever using Inverse Cloze Task (ICT), i.e., to predict its context given a sentence.
DPR~\cite{Karpukhin2020Dense} also employs two independent BERT encoders like ORQA but denies the necessity of the expensive pre-training stage. 
Instead, it focuses on learning a strong retriever using pairwise questions and answers sorely.
DPR carefully designs the ways to select negative samples to a question, including any random documents from the corpus, top documents returned by BM25 that do not contain the correct answer, and in-batch negatives which are the gold documents paired with other questions in the same batch. 
It is worth mentioning that their experiments show the inner product function is optimal for calculating the similarity score for a dual-encoder retriever.
Representation-based method~\cite{Lee2019Latent,Guu2020Realm,Karpukhin2020Dense,Karpukhin2020Dense} can be very fast since the representations of documents can be computed and indexed offline in advance.
But it may sacrifice the retrieval effectiveness because the representations of the question and document are obtained independently, leading to only shallow interactions captured between them.
    
\textbf{Interaction-based Retriever:} 
Such a kind of retrievers take a question together with a document at the same time as input, and are powerful by usually modeling the token-level interactions between them, such as transformer-based encoder~\cite{Vaswani2017Attention,Devlin2018Bert}.
\cite{Nishida2018Retrieve} propose to jointly train \emph{Retriever} and \emph{Reader} using supervised multi-task learning~\cite{Seo2017Bidirectional}.
Based on BiDAF~\cite{Seo2017Bidirectional}, a retrieval layer is added to compute the relevance score between question and document while a comprehension layer is adopted to predict the start and end position of the answer span in \cite{Nishida2018Retrieve}.
\cite{Nie2019Revealing} develop a paragraph-level dense \emph{Retriever} and a sentence-level dense \emph{Retriever}, both based on BERT~\cite{Devlin2018Bert}.
They regard the process of dense retrieval as a binary classification problem.
In particular, they take each pair of question and document as input and use the embedding of \textit{[CLS]} token to determine whether they are relevant.
Their experiments show that both paragraph-level and sentence-level retrieval are necessary for obtaining good performance of the system.
Interaction-based method~\cite{Nishida2018Retrieve, Nie2019Revealing} is powerful as it allows for very rich interactions between question and document. 
However, such a method usually requires heavy computation, which is sometimes prohibitively expensive, making it hardly applicable to large-scale documents.

 \textbf{Representation-interaction Retriever:} 
In order to achieve both high accuracy and efficiency, some recent systems~\cite{Khattab2020Relevance, Zhao2020SPARTA,zhang2020dcbert}  combine representation-based and interaction-based methods.
For instance, ColBERT-QA~\cite{Khattab2020Relevance} develops its retriever based on ColBERT~\cite{Khattab2020ColBERT}, which extends the dual-encoder architecture by performing a simple token-level interaction step over the question and document representations to calculate the similarity score.
Akin to DPR~\cite{Karpukhin2020Dense}, ColBERT-QA first encodes the question and document independently with two BERT encoders.
Formally, given a question $q$ and a document $d$, with corresponding vectors denoted as  $E_q$ (length $n$) and $E_d$ (length $m$), 
the relevance score between them is computed as follows:
\begin{equation}
\label{eq:scorer}
S_{q,d} = \sum_{i=1}^{n} \max_{j=1}^{m} E_{q_i} \cdot E_{d_j}^{T}.
\end{equation}
Then, ColBERT computes the score of each token embedding of the question over all those of the document first,  and then sums all these scores as the final relevance score between $q$ and $d$.
As another example, SPARTA~\cite{Zhao2020SPARTA} develops a neural ranker to calculate the token-level matching score using dot product between a \textit{non-contextualized encoded} (e.g., BERT word embedding) question and a \textit{contextualized encoded} (e.g., BERT encoder) document.
Concretely, given the representations of the question and document, the weight of each question token is computed with max-pooling, ReLU and log sequentially; the final relevance score is the sum of each question token weight.
The representation-interaction method is a promising approach to dense retrieval, due to its good trade-off between effectiveness and efficiency. But it still needs to be further explored.

Though effective, Dense Retriever often suffers heavy computational burden when applied to large-scale documents.
In order to speed up the computation, some works propose to compute and cache the representations of all documents offline in advance~\cite{Das2019Multistep, Feldman2019Multihop, Lee2019Latent, Guu2020Realm, Karpukhin2020Dense}.
In this way, these representations will not be changed once computed, which means the documents are encoded independently of the question, to some extent sacrificing the effectiveness of retrieval. 

\subsubsection{Iterative Retriever} 
Iterative Retriever aims to search for the relevant documents from a large collection in multiple steps given a question, which is also called Multi-step Retriever.
It has been explored extensively in the past few years~\cite{Das2019Multistep,Feldman2019Multihop,Qi2019Answering,Xiong2020Answering,Zhang2020DDRQA,Mao2020Generation,Asai2020Learning, Min2019Knowledge}, especially when answering complex questions like those requiring multi-hop reasoning~\cite{Yang2018Hotpotqa,Welbl2018Constructing}.
In order to obtain a sufficient amount of relevant documents, the search queries need to vary for different steps and be reformulated based on the context information in the previous step.
In the following, we will elaborate on Iterative Retriever based on its workflow: 
1) Document Retrieval: the IR techniques used to retrieve documents in every retrieval step; 
2) Query Reformulation: the mechanism used to generate a query for each retrieval; 
3) Retrieval Stopping Mechanism: the method to decide when to terminate the retrieval process.

\textbf{Document Retrieval:} 
We first revisit the IR techniques used to retrieve documents in every retrieval step given a query.
Some works~\cite{Qi2019Answering, Zhang2020DDRQA,Min2019Knowledge} apply Sparse Retriever iteratively, and some \cite{Das2019Multistep,Feldman2019Multihop, Xiong2020Answering, Asai2020Learning} use Dense Retriever interatively.
Among the works using Sparse Retriever, GOLDEN Retriever~\cite{Qi2019Answering} adopts BM25 retrieval, while Graph Retriever~\cite{Min2019Knowledge} and DDRQA~\cite{Zhang2020DDRQA} retrieve top $K$ documents using TF-IDF.
For those with Dense Retriever, most prior systems including Multi-step Reasoner~\cite{Das2019Multistep}, MUPPET~\cite{Feldman2019Multihop} and MDR~\cite{Xiong2020Answering} use MIPS retrieval to obtain the most semantically similar documents given a representation of the question;
Path Retriever~\cite{Asai2020Learning} develops a Recurrent Neural Network (RNN) retrieval to learn to retrieve reasoning paths for a question over a Wikipedia graph, which is built to model the relationships among paragraphs based on the Wikipedia hyperlinks and article structures.

\textbf{Query Reformulation}: 
In order to obtain a sufficient amount of relevant documents, the search queries used for each step of retrieval are usually varied and generated based on the previous query and the retrieved documents.
The generated queries take each from the two forms: 
1) explicit form, i.e. a natural language query~\cite{Qi2019Answering,Zhang2020DDRQA,Mao2020Generation}; 
and 2) implicit form, i.e. a dense representation~\cite{Das2019Multistep,Feldman2019Multihop,Xiong2020Answering}.

Some works produce a new query taking the form of natural language.
For example, GOLDEN Retriever~\cite{Qi2019Answering}  recasts the query reformulation task as an MRC task because they both take a question and some context documents as input and aim to generate a string in natural language.
Instead of pursuing an answer in MRC, the target for query reformulation is a new query that helps obtain more supporting documents in the next retrieval step. 
GAR~\cite{Mao2020Generation} develops a query expansion module using a pretrained Seq2Seq model BART~\cite{Lewis2020Bart}, which takes the initial question as input and generates new queries.
It is trained by taking various generation targets as output consisting of the answer, the sentence where the answer belongs to, and the title of a passage that contains the answer.

Some other works produce dense representations to be used for searching in a latent space.
For example, Multi-step Reasoner~\cite{Das2019Multistep} adopts a Gated Recurrent Unit (GRU)~\cite{Cho2014Learning} taking token-level hidden representations from \emph{Reader} and the question as input to generate a new query vector, which is trained using Reinforcement learning (RL) by measuring how well the answer extracted by \emph{Reader} matches the ground-truth after reading the new set of paragraphs retrieved with the new question.
MUPPET~\cite{Feldman2019Multihop} applies a bidirectional attention layer adapted from~\cite{Clark2018Simple} to a new query representation $\tilde{q}$, taking each obtained paragraph $P$ and the initial question $Q$ as input.
MDR~\cite{Xiong2020Answering} uses a pre-trained masked language model (such as RoBert) as its encoder, which encodes the concatenation of the representation of the question as well as all the previous passages as a new dense query.

Comparably, the explicit query is easily understandable and controllable to humans but is constrained by the terms in the vocabulary, 
while the implicit query is generated in a semantic space, which can get rid of the limitation of the vocabulary but lacks interpretability.

\textbf{Retrieval Stopping Mechanism}:
The iterating retrieval manner yields greater possibilities to gather more relevant passages, but the retrieval efficiency would drop dramatically along with the increasing number of iterations.
Regarding the mechanism for stopping an iterative retrieval, most existing systems choose to specify a fixed number of iterations~\cite{Das2019Multistep,Qi2019Answering,Min2019Knowledge,Zhang2020DDRQA,Xiong2020Answering} or a maximum number of retrieved documents~\cite{Feldman2019Multihop,Mao2020Generation}, which can hardly guarantee the retrieval effectiveness.
\cite{Kratzwald2018Adaptive} argue that setting a fixed number of documents to be obtained for all input questions is sub-optimal and instead they develop an Adaptive Retriever based on the Document Retriever in DrQA~\cite{Chen2017Reading}. 
They propose two methods to dynamically set the number of retrieved documents for each question, i.e. a simple threshold-based heuristic method as well as a trainable classifier using ordinal ridge regression.
Since for the questions that require arbitrary hops of reasoning, it is difficult to specify the number of iterations, 
Path Retriever~\cite{Asai2020Learning} terminates its retrieval only when the end-of-evidence token (e.g. \emph{[EOE]}) is detected by its Recurrent Retriever.
This allows it to perform adaptive retrieval steps but only obtains one document at each step.
To the best of our knowledge, it is still a critical challenge to develop an efficient iterative retriever while not sacrificing accuracy.

In addition, typical IR systems pursue two optimization targets, i.e. precision and recall.
The former computes the ratio of relevant documents returned to the total number of documents returned while the latter is the number of relevant documents returned out of the total number of relevant documents in the underlying repository.
However, for OpenQA systems, recall is much more important than precision due to the post-processing usually applied to returned documents~\cite{Andrew2004Quick}, as described below.

\subsection{Document Post-processing}
\label{sec:modern-doc-post}
Post-processing over the retrieved documents from \emph{Retriever} is often needed since the retrieved documents inevitably contain irrelevant ones, and sometimes, the number of returned documents is extremely large that overwhelms the capability of \emph{Reader}.
\emph{Document Post-processing} in the modern OpenQA architecture is similar with that in the traditional one, as introduced in Section~\ref{sec:document-processing}.
It aims at reducing the number of candidate documents and only allowing the most relevant ones to be passed to the next stage.

In the past few yeas, this module has been explored with much interest~\cite{Wang2018R3RR, Lin2018Denoising, Htut2018Training, Lee2018Ranking,Banerjee2019Careful, Wang2019Multipassage,Wang2020Answering,Wang2021Improving}.
For example, $R^3$~\cite{Wang2018R3RR} adopts a neural Passage Ranker, and trains it jointly with \emph{Reader} through Reinforcement Learning (RL).
DS-QA~\cite{Lin2018Denoising} adds a Paragraph Selector to remove the noisy ones from the retrieved documents by measuring the probability of each paragraph containing the answer among all candidate paragraphs.
\cite{Htut2018Training} explore two different passage rankers that assign scores to retrieved passages based on their likelihood of containing the answer to a given question.
One is InferSent Ranker, a forward neural network that employs InferSent sentence representations~\cite{Conneau2017Supervised}, to rank passages based on semantic similarity with the question.
The other one is Relation-Networks Ranker that adopts Relation Networks~\cite{Santoro2017Simple}, focusing on measuring word-level relevance between the question and the passages.
Their experiments show that word-level relevance matching significantly improves the retrieval performance and semantic similarity is more beneficial to the overall performance.
\cite{Lee2018Ranking} develop a Paragraph Ranker using two separate RNNs following the dual-encoder architecture.
Each pair of question-passage is fed into the Ranker to obtain their representations independently and inner product is applied to compute the relevance score.
\cite{Wang2020Answering} propose a time-aware re-ranking module that incorporates temporal information from different aspects to rank the candidate documents over temporal collections of news articles.

The focus of research on this module is learning to further re-rank the retrieved documents~\cite{Lin2018Denoising,Wang2018R3RR, Lee2018Ranking,Wang2019Multipassage}.
However, with the development of Dense Retriever, recent OpenQA systems tend to develop a trainable retriever that is capable of learning to rank and retrieving the most relevant documents simultaneously, which would  result in the absence of this module.

\subsection{Reader}
\emph{Reader} is the other core component of a modern OpenQA system and also the main feature that differentiates QA systems against other IR or IE systems, which is usually implemented as a neural MRC model. 
It is aimed at inferring the answer in response to the question from a set of ordered documents, and is more challenging compared to the original MRC task where only a piece of passage is given in most cases~\cite{Richardson2013MCTest,Hermann2015Teaching,Rajpurkar2016SQuAD,Yang2018Hotpotqa,Dua2019DROP}. 
Broadly, existing Readers can be categorised into two types: Extractive Reader that predicts an answer span from the retrieved documents, and Generative Reader that generates answers in natural language using sequence-to-sequence (Seq2Seq) models.
Most prior OpenQA systems are equipped with an Extractive Reader~\cite{Chen2017Reading, Wang2018R3RR, Lin2018Denoising,Yang2019End2end,Das2019Multistep,Feldman2019Multihop,Guu2020Realm,Karpukhin2020Dense,Min2019Knowledge}, while some recent ones develop a Generative Reader~\cite{Lewis2020Retrieval,Izacard2020Leveraging,Xiong2020Answering}.

\subsubsection{Extractive Reader} 
Extractive Reader is based on the assumption that the correct answer to a given question definitely exists in the context, and usually focuses on learning to predict the start and end position of an answer span from the retrieved documents.
The approaches can be divided into two types according to whether the retrieved documents are processed independently or jointly for answer extraction.

Many prior systems~\cite{Lin2018Denoising,Karpukhin2020Dense,Min2019Knowledge,Asai2020Learning,Zhang2020DDRQA} rank the retrieved documents by the probability of including the answer and extract the answer span from the concatenation of the question and the most probable document(s).
For example, DS-QA~\cite{Lin2018Denoising} extracts an answer span from the paragraph selected by a dedicated Paragraph Selector module through measuring the probability of each paragraph containing the answer among all candidate paragraphs.
DPR~\cite{Karpukhin2020Dense} computes the probability of a passage containing the answer and that of a token being the starting and ending position of an answer span using BERT Reader, and selects the answer with the highest probability after combining them.
Some systems develop graph-based Reader~\cite{Min2019Knowledge,Asai2020Learning} to learn to extract an answer span from a retrieved graph.
For example, Graph Reader~\cite{Min2019Knowledge} takes the graph as input and learns the passage representation mainly using Graph Convolution Networks~\cite{Kipf2017Semi} first, and then extracts the answer span from the most probable one.
Path Retriever~\cite{Asai2020Learning} leverages BERT Reader to simultaneously re-rank the reasoning paths and extract the answer span from the one with the highest probability of containing the correct answer using multi-task learning.
However, with retrieved documents processed independently, the model fails to take advantage of different evidences from the long narrative document or multiple documents for answer extraction, harming the performance especially in cases where the given questions require multiple hop reasoning.

In contrast, some systems~\cite{Chen2017Reading,Clark2018Simple,Yang2019End2end,Zhao2020SPARTA} extract an answer span based on all retrieved documents in a joint manner.
For example, DrQA~\cite{Chen2017Reading} decomposes the retrieved documents into paragraphs and extracts various features consisting of Part-of-Speech (POS), Named Entity (NE) and Term-Frequency (TF), etc.
Then the DrQA Reader, which is implemented with a multi-layer Bi-LSTM, takes as input the question and the paragraphs, and predicts an answer span.
In this process, to make answer scores comparable across paragraphs, it adopts the unnormalized exponential function and takes argmax over all answer spans to obtain the final result. % exp 
BERTserini~\cite{Yang2019End2end} develops its Reader based on BERT by removing the softmax layer to enable answer scores to be compared and aggregated among different paragraphs.
\cite{Clark2018Simple} argue using un-normalized scores (e.g., exponential scores or logits score) for all answer spans is sub-optimal and propose a Shared-Normalization mechanism by modifying the objective function to normalize the start and end scores across all paragraphs, achieving consistent performance gain.
After that, many OpenQA systems~\cite{Feldman2019Multihop,Das2019Multistep,Wang2019Multipassage,Qi2019Answering,Khattab2020Relevance,Zhao2020SPARTA} develop their readers by applying this mechanism based on original MRC models like BiDAF~\cite{Seo2017Bidirectional}, BERT~\cite{Devlin2018Bert} and SpanBERT~\cite{Joshi2019Spanbert}.

\subsubsection{Generative Reader}
Generative Reader aims to generate answers as natural as possible instead of extracting answer spans, usually relying on Seq2Seq models.
For example, S-Net~\cite{Tang2018S-Net} is developed by combining extraction and generation methods to complement each other.
It employs an evidence extraction model to predict the boundary of the text span as the evidence to the answer first and then feeds it into a Seq2Seq answer synthesis model to generate the final answer.
Recently, some OpenQA systems~\cite{Lewis2020Retrieval,Izacard2020Leveraging,Xiong2020Answering} adopt pretrained Seq2Seq language models to develop their Readers, like BART~\cite{Lewis2020Bart} and T5~\cite{Raffel2019T5}. 
For example, RAG~\cite{Lewis2020Retrieval} adopts a pretrained BART model as its reader to generate answers by taking the input question as well as the documents retrieved by DPR~\cite{Karpukhin2020Dense}.
FID~\cite{Izacard2020Leveraging} first encodes each retrieved document independently using T5 or BART encoder and then performs attention over all the output representations using the decoder to generate the final answer.
However, the current generation results often suffer syntax error, incoherence, or illogic~\cite{Liu2019Neural}.
Generative Reader needs to be further explored and advanced.

\subsection{Answer Post-processing}
Neural MRC techniques have been advancing rapidly in recent years, but most existing MRC models still specialise in extracting answers only from a single or several short passages and tend to fail in cases where the correct answer comes from various evidences in a narrative document or multiple documents~\cite{Wang2018Evidence}.
The \emph{Answer Post-processing} module is developed to help detect the final answer from a set of answer candidates extracted by the \emph{Reader}, taking into account their respective supporting facts.
The methods adopted in existing systems can be classified into two categories, i.e. rule-based method~\cite{Wang2018Evidence,Lee2018Ranking} and learning-based method~\cite{Wang2018Evidence,Kratzwald2019RankQA} 
For example, \cite{Wang2018Evidence} propose two answer re-rankers, a ``strength-based re-ranker'' and a ``coverage-based re-ranker'', to aggregate evidences from different passages to decide the final answer.
The ``strength-based re-ranker'' is a rule-based method that simply performs counting or probability calculation based on the candidate predictions and does not require any training.
The ``coverage-based re-ranker''  is developed using an attention-based match-LSTM model~\cite{Wang2016Learning}.
It first concatenates the passages containing the same answer into a pseudo passage and then measures how well this pseudo passage entails the answer for the given question.
The experiments in \cite{Wang2018Evidence} show that a weighted combination of the outputs of the above different re-rankers achieves the best performance on several benchmarks.
RankQA~\cite{Kratzwald2019RankQA} develops an answer re-ranking module consisting of three steps: feature extraction, answer aggregation and re-ranking.
Firstly, taking the top $k$ answer candidates from \emph{Reader} as input, the module extracts a set of features from both \emph{Retriever} such as document-question similarity, question length and paragraph length,  and \emph{Reader} such as original score of answer candidate, part-of-speech tags of the answer and named entity of the answer.
Secondly, the module groups all answer candidates with an identical answer span to generate a set of aggregation features like the sum, mean, minimum, and maximum of the span scores, etc.
Based on these features, a re-ranking network such as a feed-forward network or an RNN is used to learn to further re-rank the answers to select the best one as the final answer.

\subsection{End-to-end Methods}
In recent years, various OpenQA systems~\cite{Nishida2018Retrieve,Lee2019Latent,Guu2020Realm} have been developed, in which \emph{Retriever} and \emph{Reader} can be trained in an end-to-end manner.
In addition, there are some systems with only Retriever~\cite{Seo2019Real}, and also some that are able to answer open questions without the stage of retrieval, which are mostly pre-trained Seq2Seq language models~\cite{Raffel2019T5, Lewis2020Bart,radford2019GPT2,Brown2020GPT3}.
In the following, we will introduce these three types of systems, i.e. \textit{Retriever-Reader}, \textit{Retriever-only} and \textit{Retriever-free}.

\subsubsection{Retriever-Reader}
Deep learning techniques enable \emph{Retriever} and \emph{Reader} in an OpenQA system to be end-to-end trainable~\cite{Nishida2018Retrieve,Lee2019Latent,Guu2020Realm,Lewis2020Retrieval}.
For example, \cite{Nishida2018Retrieve} propose to jointly train \emph{Retriever} and \emph{Reader} using  multi-task learning based on the BiDAF model~\cite{Seo2017Bidirectional}, simultaneously computing the similarity of a passage to the given question and predicting the start and end position of an answer span.
\cite{Lee2019Latent} argue that it is sub-optimal to incorporate a standalone IR system in an OpenQA system and develop ORQA that jointly trains \emph{Retriever} and \emph{Reader} from question-answer pairs, with both developed using BERT~\cite{Devlin2018Bert}.
REALM~\cite{Guu2020Realm} is a pre-trained masked language model including a neural \emph{Retriever} and a neural \emph{Reader}, which is able to compute the gradient w.r.t. the model parameters and backpropagate the gradient all the way throughout the network.
Since both modules are developed using neural networks, the response speed to a question is a most critical issue during inference, especially over a large collection of documents.

\subsubsection{Retriever-only}
To enhance the efficiency of answering questions, some systems are developed by only adopting a \emph{Retriever} while omitting \emph{Reader} which is usually the most time-consuming stage in other modern OpenQA systems.
DenSPI~\cite{Seo2019Real} builds a question-agnostic phrase-level embedding index offline given a collection of documents like Wikipedia articles.
In the index, each candidate phrase from the corpus is represented by the concatenation of two vectors, i.e. a sparse vector (e.g., tf-idf) and a dense vector (e.g., BERT encoder).
In the inference, the given question is encoded in the same way, and FAISS~\cite{Jeff2017Billion} is employed to search for the most similar phrase as the final answer.
Experiments show that it obtains remarkable efficiency gains and reduces computational cost significantly while maintaining accuracy.
However, the system computes the similarity between each phrase and the question independently, which ignores the contextual information that is usually crucial to answering questions .

\subsubsection{Retriever-free}
\label{sec:retriever-free}
Recent advancement in pre-training Seq2Seq language models such as GPT-2~\cite{radford2019GPT2}, GPT-3~\cite{Brown2020GPT3}, BART~\cite{Lewis2020Bart} and T5~\cite{Raffel2019T5} brings a surge of improvements for downstream NLG tasks, most of which are built using Transformer-based architectures.
In particular, GPT-2 and GPT-3 adopt Transformer left-to-right decoder while BART and T5 use Transformer encode-decoder closely following its original form~\cite{Vaswani2017Attention}.
Prior studies~\cite{Petroni2020Language,Roberts2020How} show that a large amount of knowledge learned from large-scale textual data can be stored in the underlying parameters, and thus these models are capable of answering questions without access to any external knowledge.
For example, GPT-2~\cite{radford2019GPT2} is able to correctly generate the answer given only a natural language question without fine-tuning.
Afterwards, GPT-3~\cite{Brown2020GPT3} achieves competitive performance with few-shot learning compared to prior state-of-the-art fine-tuning approaches, in which several demonstrations are given at inference as conditioning~\cite{radford2019GPT2} while weight update is not allowed. 
Recently, \cite{Roberts2020How} comprehensively evaluate the capability of language models for answering questions without access to any external knowledge.
Their experiments demonstrate that pre-trained language models are able to gain impressive performance on various benchmarks and such Retrieval-free methods make a fundamentally different approach to building OpenQA systems.

In Table~\ref{tab:qa-system}, we summarize existing modern OpenQA systems as well as the approaches adopted for different components.

\begin{ThreePartTable}{
\begin{table*}[!ht]
 \caption{Approaches adopted for different components of existing modern OpenQA systems.}
  \small{
    \centering
    \begin{tabular}{llllll@{}}
    \toprule
    \bf System  & \bf Category & \makecell{\bf Retriever}  & \bf Document \bf Post -Processing & \makecell[l]{\bf Reader } &  \bf Answer \bf Post-processing \\
    \toprule
    DrQA~\cite{Chen2017Reading} & Pipeline & Sparse & - & Extractive & - \\
    R3~\cite{Wang2018R3RR} & Pipeline & Sparse & RL\tnote{1} & Extractive & - \\
    DS-QA~\cite{Lin2018Denoising} & Pipeline & Sparse & SL\tnote{2} & Extractive & - \\
    \cite{Wang2018Evidence} & Pipeline & Sparse & - & Extractive & \makecell[l]{Rule-based \\  Learning-based} \\
    \cite{Htut2018Training} & Pipeline & Sparse & SL & Extractive & - \\
    \makecell[l]{Paragraph \\ Ranker~\cite{Lee2018Ranking}} & Pipeline & Sparse & SL & Extractive & Rule-based \\
    RankQA~\cite{Kratzwald2019RankQA} & Pipeline & Sparse & - & Extractive & Learning-based \\
    BERTserini~\cite{Yang2019End2end} & Pipeline & Sparse & - & Extractive & - \\
	\makecell[l]{Multi-Passage \\ BERT~\cite{Wang2019Multipassage}} & Pipeline & Sparse & TL\tnote{3} & Extractive & - \\
    \cite{Nishida2018Retrieve} & Pipeline & Dense & - & Extractive & - \\
    DPR~\cite{Karpukhin2020Dense} & Pipeline & Dense & - & Extractive & - \\
    ColBERT-QA~\cite{Khattab2020Relevance} & Pipeline & Dense & - & Extractive & - \\
    SPARTA~\cite{Zhao2020SPARTA} & Pipeline & Dense & - & Extractive & - \\
    \makecell[l]{FID~\cite{Izacard2020Leveraging}} & Pipeline &\makecell[l]{Sparse \\ Dense} & - & Generative & - \\
    \makecell[l]{Adaptive \\ Retrieval \cite{Kratzwald2018Adaptive}} & Pipeline & Iterative & - & Extractive & - \\
    \makecell[l]{Multi-step \\ Reasoner~\cite{Das2019Multistep}} & Pipeline & Iterative & - & Extractive & - \\
    \makecell[l]{GOLDEN \\ Retriever~\cite{Qi2019Answering}} & Pipeline & Iterative & - & Extractive & - \\
    MUPPET~\cite{Feldman2019Multihop} & Pipeline & Iterative & - & Extractive & - \\
    Path Retriever~\cite{Asai2020Learning} & Pipeline & Iterative & - & Extractive & - \\
    \makecell[l]{Graph \\Retriever~\cite{Min2019Knowledge}} & Pipeline & Iterative & - & Extractive & - \\
    DDRQA~\cite{Zhang2020DDRQA} & Pipeline & Iterative & - & Extractive & - \\
    GAR~\cite{Mao2020Generation} & Pipeline & Iterative & - & Extractive & Rule-based \\
    MDR~\cite{Xiong2020Answering} & Pipeline & Iterative & - & \makecell{Extractive \\ Generative} & - \\
    \midrule
    DenSPI~\cite{Seo2019Real} & End-to-end & Retriever & - & - & - \\
    \makecell[l]{Retrieve-\\-and-Read~\cite{Nie2019Revealing}} & End-to-end & Dense & - & Extractive & - \\
    ORQA~\cite{Lee2019Latent} & End-to-end & Dense & - & Extractive & - \\
    REALM~\cite{Guu2020Realm} & End-to-end & Dense & - & Extractive & - \\
    RAG~\cite{Lewis2020Retrieval} & End-to-end & Dense & - & Generative & - \\
    \bottomrule
    \end{tabular}
     \begin{tablenotes}
        \scriptsize
             \item [1] RL: Reinforcement Learning;
             \item [2] SL: Supervised Learning;
             \item [3] TL: Transfer Learning.
       \end{tablenotes}
    \label{tab:qa-system}
    }
\end{table*}
}
\end{ThreePartTable}

%% file: 04_Challenge.tex
\label{challengesandbenchmarks}

In this section, we first discuss key challenges to building OpenQA systems followed by an analysis of existing QA benchmarks that are commonly used not only for OpenQA but also for MRC.

\subsection{Challenges to OpenQA}
To build an OpenQA system that is capable of answering any input questions is regarded as the ultimate goal of QA research.
However, the research community still has a long way to go. 
Here we discuss some salient challenges that need be addressed on the way.
By doing this we hope the research gaps can be made clearer so as to accelerate the progress in this field.

\subsubsection{Distant Supervision}

In the OpenQA setting, it is almost impossible to create a collection containing ``sufficient'' high-quality training data for developing OpenQA systems in advance.
Distant supervision is therefore popularly utilized, which is able to label data automatically based on an existing corpus, such as Wikipedia.
However, distant supervision inevitably suffers from wrong label problem and often leads to a considerable amount of noisy data, significantly increasing the difficulty of modeling and training.
Therefore, the systems that are able to tolerate such noise are always demanded.

\subsubsection{Retrieval Effectiveness and Efficiency}
Retrieval effectiveness means the ability of the system to separate relevant documents from irrelevant ones for a given question.
The system often suffers from ``term-mismatch'', which results in failure of retrieving relevant documents;
on the other hand the system may receive noisy documents that contain the exact terms in the question or even the correct answer span, but are irrelevant to the question.
Both issues increase the difficulty of accurately understanding the context during answer inference.
Some neural retrieval methods~\cite{Seo2018Phrase, Nishida2018Retrieve,Seo2019Real,Lee2019Latent,Guu2020Realm, Karpukhin2020Dense} are proposed recently for improving retrieval effectiveness. 
For example, \cite{Lee2019Latent} and \cite{Guu2020Realm} jointly train the retrieval and reader modules, which take advantage of pre-trained language models and regard the retrieval model as a latent variable. 
However, these neural retrieval methods often suffer from low efficiency. 
Some works~\cite{Nishida2018Retrieve, Lee2019Latent, Seo2018Phrase, Karpukhin2020Dense} propose to pre-compute the question-independent embedding for each document or phrase and construct the embedding index only once.
Advanced sub-linear \emph{Maximum Inner Product Search (MIPS)} algorithms~\cite{Ram2012Maximum,Shrivastava2014ALSH,Shen2015Learning} are usually employed to obtain the top $K$ related documents given a question.
However, the response speed still has a huge gap from that of typical IR techniques when the system faces a massive set of documents.

Retrieval effectiveness and efficiency are both crucial factors for the deployment of an OpenQA system in practice, especially when it comes to the real-time scenarios. 
How to consistently enhance both aspects (also with a good trade-off) will be a long-standing challenge in the advancement of OpenQA.

\subsubsection{Knowledge Incorporation}
To incorporate knowledge beyond context documents and given questions is a key enhancement to OpenQA systems~\cite{Burger2001Issues}, e.g. world knowledge, commonsense or domain-specific knowledge.
Before making use of such knowledge, we need to first consider how to represent them.
There are generally two ways: \texttt{explicit} and \texttt{implicit}.

For the \texttt{explicit} manner, knowledge is usually transformed into the form of triplets and stored in classical KBs such as DBPedia~\cite{Auer2007DBpedia}, Freebase~\cite{Bollacker2008Freebase} and  Yago2~\cite{Hoffart2013YAGO2}, which are easily understood by humans.
Some early QA systems attempt to incorporate knowledge to help find the answer in this way.
For example, IBM Watson DeepQA~\cite{Ferrucci2010Watson} combines a Web search engine and a KB to compete with human champions on the American TV show ``Jeopardy''; 
QuASE~\cite{Sun2015Open} searches for a list of most prominent sentences from a Web search engine (e.g, Google.com), and then utilizes entity linking over Freebase~\cite{Bollacker2008Freebase} to detect the correct answer from the selected sentences.
In recent years, with the popularity of Graph Neural Network (GNN), some works \cite{Sun2018Open, Sun2019Pullnet, Min2019Knowledge} propose to gain relevant information not only from a text corpus but also from a KB to facilitate evidence retrieval and question answering.
For example, \cite{Sun2018Open} construct a question-specific sub-graph containing sentences from the corpus, and entities and relations from the KB.
Then, graph CNN based methods~\cite{Kipf2017Semi, Li2016Gated,Scarselli2009GNN}  are used to infer the final answer over the sub-graph.
However, there also exist problems for storing knowledge in an explicit manner, such as incomplete and out-of-date knowledge.
Moreover, to construct a KB is both labor-intensive and time-consuming.

On the other hand,
with the \texttt{implicit} approach, a large amount of knowledge~\cite{Petroni2020Language} can be stored in underlying parameters learned from massive texts by pre-trained language models such as BERT~\cite{Devlin2018Bert}, XLNet~\cite{Dai2019XLNet} and T5~\cite{Raffel2019T5}, which can be applied smoothly in downstream tasks.
Recently, pre-trained language models have been popularly researched and applied to developing OpenQA systems~\cite{Yang2019End2end, Lee2019Latent, Nie2019Revealing, Asai2020Learning, Guu2020Realm, Karpukhin2020Dense, Mao2020Generation}.
For example, \cite{Yang2019End2end, Nie2019Revealing, Asai2020Learning} develop their \emph{Reader} using BERT~\cite{Devlin2018Bert} while 
\cite{Lee2019Latent, Karpukhin2020Dense} use BERT to develop both~\emph{Retriever} and~\emph{Reader}.
In addition, pre-trained language models like GPT-2~\cite{radford2019GPT2} are able to generate the answer given only a natural language question.
However, such systems act like a ``black box'' and it is nearly impossible to know what knowledge has been exactly stored and used for a particular answer.
They lack interpretability that is crucial especially for real-world applications.

Knowledge enhanced OpenQA is desired not only because it is helpful to generating the answer but also because it serves as the source for interpreting the obtained answer.
How to represent and make full use of the knowledge for OpenQA still needs more research efforts.
% \\~

\begin{ThreePartTable}

\begin{table*}[htb]
\caption{\textbf{Dataset:} The name of the dataset. 
        \textbf{Domain:} The domain of background information in the dataset.
        \textbf{\#Q (k):} The number of questions contained in the dataset, with unit(k) denoting ``thousand''.
        \textbf{Answer Type:} The answer types included in the dataset.
        \textbf{Context in MRC:} The context documents or passages that are given to generate answers in MRC tasks.
        \textbf{OpenQA:} This column indicates whether the dataset is applicable for developing OpenQA systems, with the tick mark denoting yes.
        }
        \centering
    \scriptsize{
 	
    \begin{tabular}{llrlp{0.35\textwidth}c}
        \\
        \toprule
        \bf Dataset   & \bf Domain & \bf \makecell[lt]{\#Q (k)} & \bf Answer Type & \bf Context  in MRC  & \bf \makecell{OpenQA}\\
        \midrule
        MCTest~\cite{Richardson2013MCTest}   & Children's story & 2.0 & Multiple choices & A children's story & \\
        CNN/Daily Mail~\cite{Hermann2015Teaching}   & News & 1,384.8 & Entities &
        A passage from one CNN or Daily Mail news & \ding{52} \\
        CBT~\cite{Hill2015CBT}  & Children's story & 687.3 &
        Multiple choices & A children's story & \\
        SQuAD~\cite{Rajpurkar2016SQuAD}& Wikipedia & 108.0 & Spans & A passage from Wikipedia & \ding{52} \\
        MS MARCO~\cite{Nguyen2016MS}   & Web search & 1,010.9 &
        \makecell[lt]{Free-form\\Boolean\\Unanswerable} & Multiple passages from Bing Search & \ding{52} \\
        NewsQA~\cite{Trischler2017Newsqa}   & News & 119.6 &
        \makecell[lt]{Spans\\Unanswerable} & A news article from CNN news & \ding{52} \\
        SearchQA~\cite{Dunn2017SearchQA}   & Web search & 140.4 & Spans & Multiple passages from Google Search & \ding{52} \\
        TriviaQA~\cite{Joshi2017Triviaqa}   & Trivia & 95.9 &
        \makecell[lt]{Spans\\Free-form} & One or multiple passages & \\
        RACE~\cite{Lai2017RACE}   & Science & 97.6 &Multiple choices & A passage from mid/high school exams & \\
        Quasar-T~\cite{Welbl2018Constructing}   & Reddit & 43.0 & Free-form & Multiple documents from Reddit & \ding{52}\\
        Quasar-S~\cite{Welbl2018Constructing}  & Technical & 37.0 & Entities & A passage from Stack Overflow & \ding{52}\\
        NarrativeQA~\cite{Tom2017NarrativeQA}   & Others & 46.7 & Free-form &
        A summary and a full story from movie scripts &  \\
        DuReader~\cite{He2017DuReader}   & Web search & 200.0 &
        \makecell[lt]{Free-form\\Boolean} & Multiple passages from Baidu Search or Baidu Zhidao & \ding{52} \\
        SQuAD 2.0~\cite{Rajpurkar2018Know}  & Wikipedia & 158.0 &
        \makecell[lt]{Spans\\Unanswerable} & A passage from Wikipedia & \ding{52} \\
        CoQA~\cite{Reddy2018CoQA}   & Others & 127.0 &
        \makecell[lt]{Free-form\\Boolean\\Unanswerable} & A passage and conversation history & \\
        QuAC~\cite{Choi2018QuAC}  & Wikipedia & 98.4 &
        \makecell[lt]{Spans\\Boolean\\Unanswerable} & A passage from Wikipedia and conversation history & \ding{52} \\
        ARC~\cite{Clark2018Think}   & Science & 7.7 &Multiple choices & No additional context & \\
        ShARC~\cite{Saeidi2018ShARC}   & Others & 32.4 & Boolean & A rule text, a scenario and conversation history & \\
        CliCR~\cite{Suster2018Clicr}   & Medical & 104.9 & Spans & A passage from clinical case reports & \\
        HotpotQA~\cite{Yang2018Hotpotqa}  & Wikipedia & 113.0 &
        \makecell[lt]{Spans\\Boolean\\Unanswerable} & A pair of paragraphs from Wikipedia & \ding{52} \\
        MultiRC~\cite{Khashabi2018Looking} & Others & 6.0 & Multiple choices & Multiple sentences & \\
        SWAG~\cite{Zellers2018SWAG}  & Commonsense & 113.0 & Multiple choices & A piece of video caption & \\
        DuoRC~\cite{Amrita2018DuoRC} & Others & 186.0 &
        \makecell[lt]{Free-form\\Spans\\Unanswerable} & A movie plot story & \\
        WikiHop~\cite{Welbl2018Constructing}& Wikipedia & 51.3 & Multiple choices & Multiple passages from Wikipedia & \ding{52} \\
        MedHop~\cite{Welbl2018Constructing}  & Medical & 2.5 & Multiple choices & Multiple passagee from MEDLINE & \\
        ReCoRD~\cite{Zhang2018ReCoRD} & News & 120.7 & Multiple choices &
        A passage from CNN/Daily Mail News & \\
        OpenBookQA~\cite{Todor2018OpenBookQA}  & Science & 5.9 & Multiple choices & Open book &  \\
        CommonsenseQA~\cite{Talmor2018CommonsenseQA}  & Commonsense & 12.2 & Multiple choices & No additional context & \\
        CODAH~\cite{Chen2019CODAH}  & Commonsense & 2.8 & Multiple choices & No additional context & \\
        DROP~\cite{Dua2019DROP}   & Wikipedia & 96.5 &
        Free-form & A passage from Wikipedia & \\
        \makecell[l]{Natural \\ Questions~\cite{Kwiatkowski2019Natural}}   & Wikipedia & 323.0 &
        \makecell[lt]{Spans\\Boolean\\Unanswerable} & An article from Wikipedia & \ding{52} \\
        Cosmos QA~\cite{Huang2019CosmosQA} & Commonsense & 35.6 & Multiple choices & A passage & \\
        BoolQ~\cite{Clark2019Boolq}   & Wikipedia & 16.0 & Boolean & An article from Wikipedia & \ding{52} \\
        ELI5~\cite{Fan2019Eli5}   & Reddit & 272.0 & Free-form & A set of web documents & \\
        TWEETQA~\cite{Xiong2019Tweetqa}   & Social media & 13.7 & Free-form & A tweet from Twitter & \\
        XQA~\cite{Liu2019XQA}   & Wikipedia & 90.6 & Entities & A passage from Wikipedia in a target language & \ding{52} \\
        \bottomrule
\end{tabular}
}
\label{tab:benchmark}
\end{table*}
\end{ThreePartTable}

\subsubsection{Conversational OpenQA}
Non-conversational OpenQA is challenged by several problems that are almost impossible to resolve, such as the lengthy words for a complex question (e.g. \textsl{Who is the second son of the first Prime Minister of Singapore?}), ambiguity resulting in incorrect response (e.g. \textsl{When was Michael Jordan born?})
and insufficient background knowledge from the user that leads to unreasonable results (e.g. \textsl{Why do I have a bad headache today?}).
These problems would be well addressed under the conversational setting.

Conversational systems~\cite{Gao2019Neural,Lei2020Conversational} are equipped with a dialogue-like interface that enables interaction between human users and the system for information exchange.
For the complex question example given above, it can be decomposed into two simple questions sequentially: \textsl{``Who is the first Prime Minister of Singapore?''} followed by \textsl{``Who is the second son of him?''}.
When ambiguity is detected in the question, the conversational OpenQA system is expected to raise a follow-up question for clarification, such as \textsl{``Do you mean the basketball player?''}.
If a question with insufficient background knowledge is given, a follow-up question can also be asked to gather more information from human users for arriving at the final answer.
To achieve these goals, three major challenges need to be addressed.

First, conversational OpenQA should have the ability to determine if a question is unanswerable, such as to detect if ambiguity exists in the question or whether the current context is sufficient for generating an answer.
Research on unanswerable questions has attracted a lot of attention in the development of MRC over the past few years~\cite{Rajpurkar2018Know, Nguyen2016MS, Kwiatkowski2019Natural, Trischler2017Newsqa, Zhu2019Learning, Hu2018ReadV}.
However, current OpenQA systems rarely incorporate such a mechanism to determine unanswerability of questions, which is particularly necessary for conversational OpenQA systems.

Second, when the question is classified as unanswerable due to ambiguity or insufficient background knowledge, the conversational OpenQA system needs to generate a follow-up question~\cite{Aliannejadi2019Asking}.
Question Generation (QG) can then be considered as a sub-task of QA,  which is a crucial module of conversational OpenQA.
In the past few years, research on automatic question generation from text passages has received growing attention~\cite{Du2017Learning, Duan2017Question,Zhou2017Neural,Pan2019Recent}.
Compared to the typical QG task targeting at generating a question based on a given passage where the answer to the generated question can be found, the question generated in conversational OpenQA should be answered by human users only.

The third challenge is how to better model the conversation history not only in \emph{Reader} but also in \emph{Retriever}~\cite{Chen2020Open}. 
The recently released conversational MRC datasets like CoQA~\cite{Reddy2018CoQA} and QuAC~\cite{Choi2018QuAC} are aimed at enabling a \emph{Reader} to answer the latest question by comprehending not only the given context passage but also the conversation history so far.
As they provide context passages in their task setting, they omit the stage of document retrieval which is necessary when it comes to OpenQA.
Recently, in \cite{Chen2020Open} the QuAC dataset is extended to a new OR-QuAC dataset by adapting to an open-retrieval setting, and an open-retrieval conversational question answering system (OpenConvQA) is developed, which is able to retrieve relevant passages from a large collection before inferring the answer, taking into account the conversation QA pairs.
OpenConvQA tries to answer a given question without any specified context, and thus enjoys a wider scope of application and better accords with real-world QA behavior of human beings.
However, the best performance (F1: 29.4) of the system on OR-QuAC is far lower than the state-of-the-art (F1: 74.4\footnote{stated on June 2020 https://quac.ai/}) on QuAC, indicating that it is a bigger challenge when it comes to an open-retrieval setting.

\begin{figure*}[ht]
  \centering
  \caption{Number of questions in each dataset}
  \includegraphics[width=0.9\textwidth]{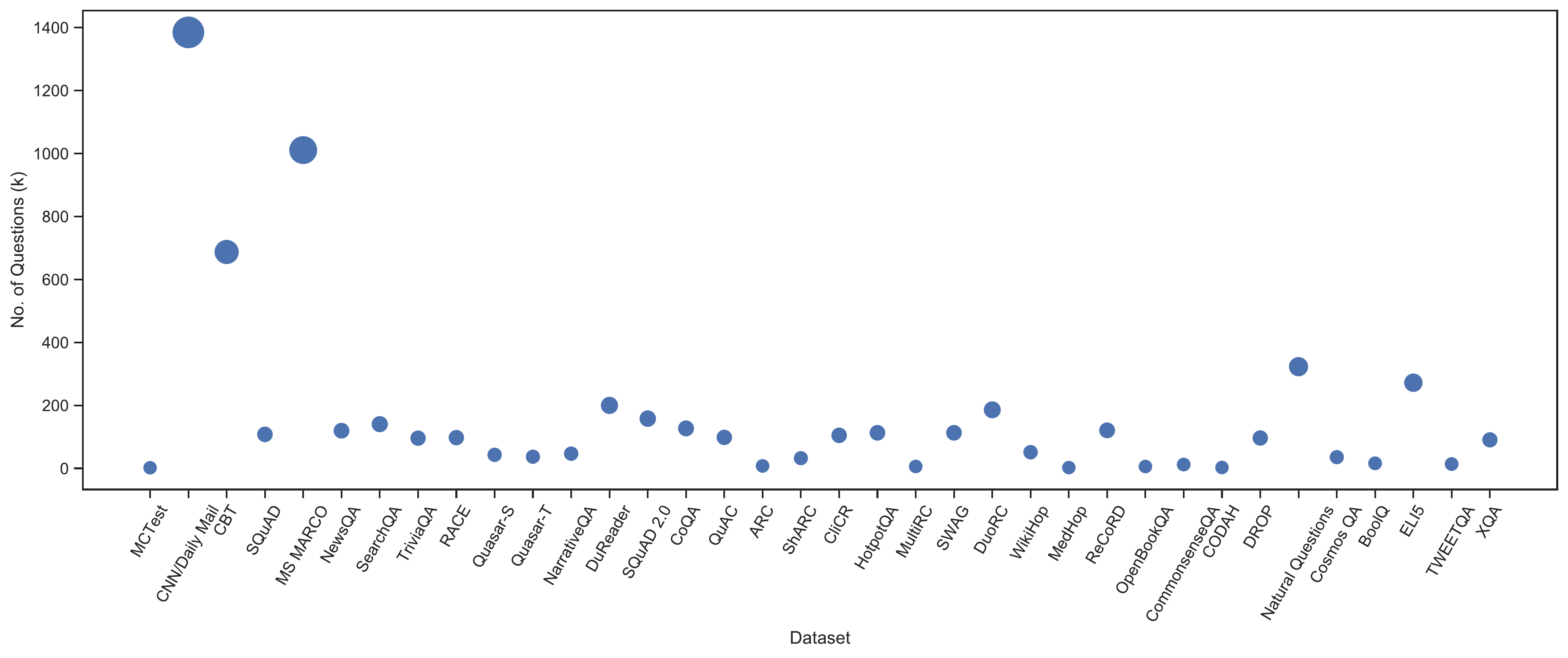}
  \label{fig:question_rel}
\end{figure*}

\begin{figure}[ht]
  \centering
  \caption{Distribution of popular datasets w.r.t. release year}
    \includegraphics[width=0.9\linewidth]{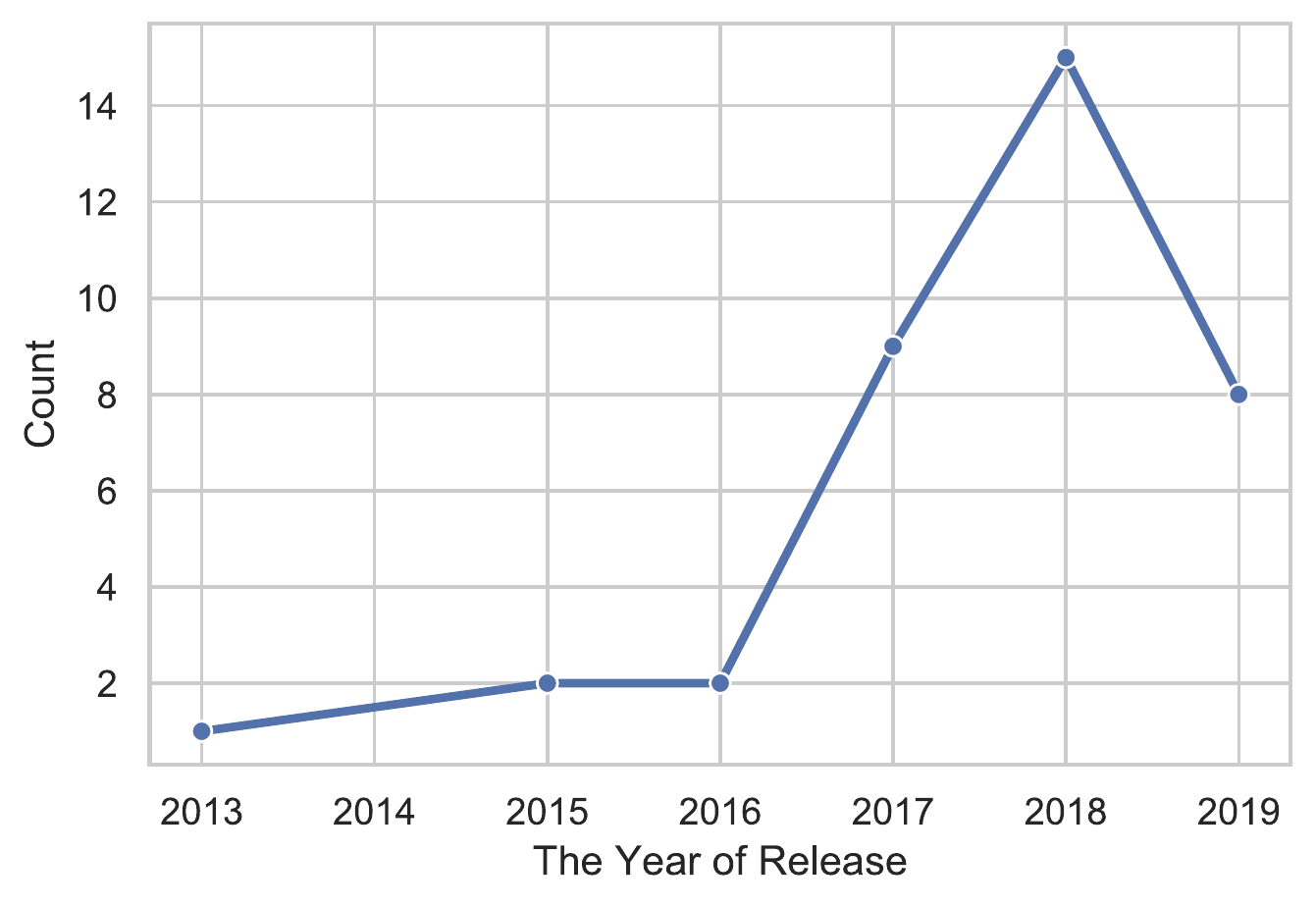}
      \label{fig:release_year}
\end{figure}

\begin{figure}[ht]
  \centering
  \caption{Datasets distribution w.r.t. background information domain}
    \includegraphics[width=0.9\linewidth]{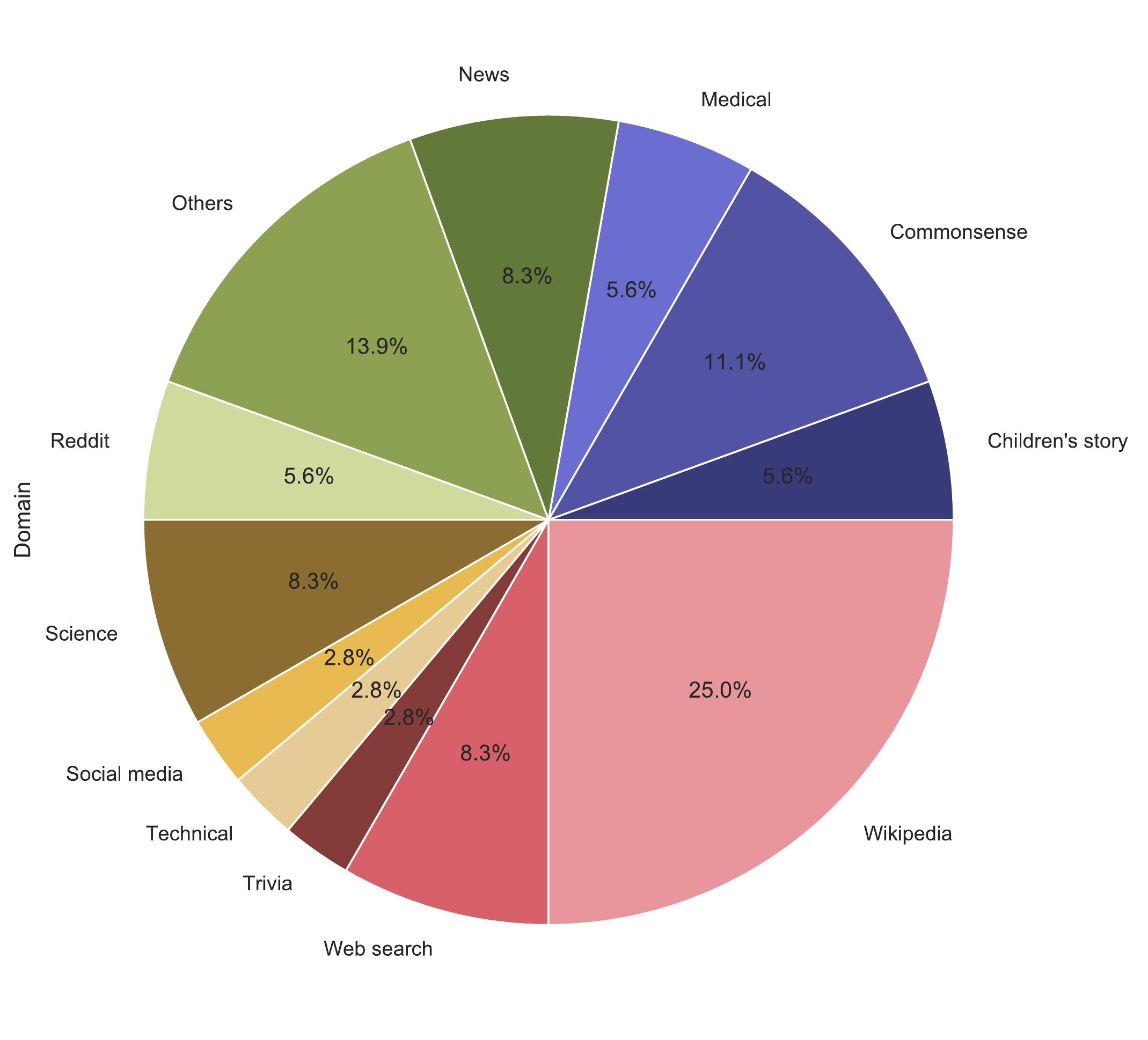}
      \label{fig:domain}
\end{figure}

\begin{ThreePartTable}
    \begin{TableNotes}
        \small
    \end{TableNotes}
    \begin{table*}[ht] %%%%% here to adjust the horizontal 
        \caption{The information source of the datasets that are applicable for developing OpenQA system. 
        \textbf{Source Type:} The type of background information source. 
        \textbf{Source}: The background information source in OpenQA setting. 
        }
        \label{tab:openqa}
        \centering
        
        \begin{tabular}{ll|l} \toprule
         \bf Source Type & \bf  Source  & \bf Dataset \\
         \midrule
      \multirow{7}{*}{ Wikipedia } & \multirow{7}{*}{ Full Wikipedia } & $SQuAD_{open}$~\cite{Chen2017Reading} \\
        & & HotpotQA~\cite{Yang2018Hotpotqa} \\
        & & QuAC~\cite{Choi2018QuAC} \\
        & & WikiHop~\cite{Welbl2018Constructing} \\
        & & Natural Questions~\cite{Kwiatkowski2019Natural} \\
        & & BoolQ~\cite{Clark2019Boolq} \\
        & & XQA~\cite{Liu2019XQA} \\
          \midrule
        \multirow{3}{*}{ Search Engine } & Bing Search & MS MARCO~\cite{Nguyen2016MS} \\
        & Google Search & SearchQA~\cite{Dunn2017SearchQA} \\
        & Baidu Search & DuReader~\cite{He2017DuReader} \\
          \midrule
        \multirow{2}{*}{ Online News } & News from CNN/Daily & CNN/Daily Mail~\cite{Hermann2015Teaching} \\
        & News from CNN & NewsQA~\cite{Trischler2017Newsqa} \\
          \midrule
        \multirow{2}{*}{ Internet Forum } & Reddit & Quasar-T~\cite{Welbl2018Constructing} \\
        & Stack Overflow & Quasar-S~\cite{Welbl2018Constructing} \\

        \bottomrule
        \end{tabular}\\
 
    \end{table*}
\end{ThreePartTable}
\subsection{Benchmarks}
A large number of QA benchmarks have been released in the past decade, which are summarized in Table~\ref{tab:benchmark}. 
Here we provide a brief analysis of them with the focus on their respective characteristics, dataset distributions w.r.t. background information domain, number of questions, year of release.
As aforementioned in this paper, the success of the MRC task is a crucial step to more advanced OpenQA and we believe the future advancement of MRC methods will significantly promote the OpenQA systems.
Thus, we include not only the datasets for OpenQA but also those solely for MRC to make our survey more comprehensive.

The major criterion for judging the applicability of a QA dataset to develop OpenQA systems is whether it involves a separate document set (usually large-scale) \cite{Yang2018Hotpotqa}, or whether it has relatively easy access to such an information source~\cite{Rajpurkar2018Know, Hermann2015Teaching} where the answers to questions can be inferred.
For example, HotpotQA~\cite{Yang2018Hotpotqa} provides a full-wiki setting itself to require a system to find the answer to a question in the scope of the entire Wikipedia.
\cite{Chen2017Reading} extend SQuAD~\cite{Rajpurkar2016SQuAD} to $SQuAD_{open}$ by using the entire Wikipedia as its information source.
We summarize and illustrate the distributions of datasets listed in Table \ref{tab:benchmark} w.r.t. year of release in Fig. \ref{fig:release_year}, background information domain in Fig.\ref{fig:domain} and number of questions in Fig. \ref{fig:question_rel}.
Also, we summarize the information source type of the datasets that are applicable to developing OpenQA systems in Table \ref{tab:openqa}.

%% file: 05_Conclusion.tex
\label{conclusion}

In this work we presented a comprehensive survey on the latest progress of Open-domain QA (OpenQA) systems.
In particular, we first reviewed the development of OpenQA and illustrated a ``Retriever-Reader'' architecture.
Moreover, we reviewed a variety of existing OpenQA systems as well as their different approaches.
Finally, we discussed some salient challenges towards OpenQA followed by a summary of various QA benchmarks, hoping to reveal the research gaps so as to push further progress in this field.
Based on our review of prior research, we claim that OpenQA would continue to be a research hot-spot. 
In particular, single-step and multi-step neural retrievers will attract increasing attention due to the demand for more accurate retrieval of related documents.
Also, more end-to-end OpenQA systems will be developed with the advancement of deep learning techniques.
Knowledge enhanced OpenQA is very promising not only because it is helpful to generating the answer but also because it serves as the source for interpreting the obtained answer. 
However, how to represent and make full use of the knowledge for OpenQA still needs more research efforts.
Furthermore, to equip OpenQA with a dialogue-like interface that enables interaction between human users and the system for information exchange is expected to attract increasing attention, which well aligns with real world application scenarios.

%% file: main.bbl
\begin{thebibliography}{100}
\providecommand{\url}[1]{#1}
\csname url@samestyle\endcsname
\providecommand{\newblock}{\relax}
\providecommand{\bibinfo}[2]{#2}
\providecommand{\BIBentrySTDinterwordspacing}{\spaceskip=0pt\relax}
\providecommand{\BIBentryALTinterwordstretchfactor}{4}
\providecommand{\BIBentryALTinterwordspacing}{\spaceskip=\fontdimen2\font plus
\BIBentryALTinterwordstretchfactor\fontdimen3\font minus
  \fontdimen4\font\relax}
\providecommand{\BIBforeignlanguage}[2]{{%
\expandafter\ifx\csname l@#1\endcsname\relax
\typeout{** WARNING: IEEEtran.bst: No hyphenation pattern has been}%
\typeout{** loaded for the language `#1'. Using the pattern for}%
\typeout{** the default language instead.}%
\else
\language=\csname l@#1\endcsname
\fi
#2}}
\providecommand{\BIBdecl}{\relax}
\BIBdecl

\bibitem{Green1961Baseball}
B.~F. Green, Jr., A.~K. Wolf, C.~Chomsky, and K.~Laughery, ``Baseball: An
  automatic question-answerer,'' in \emph{Papers Presented at the May 9-11,
  1961, Western Joint IRE-AIEE-ACM Computer Conference}.\hskip 1em plus 0.5em
  minus 0.4em\relax ACM, 1961, pp. 219--224.

\bibitem{JOEL2011Google}
\BIBentryALTinterwordspacing
J.~Falconer, ``Google: Our new search strategy is to compute answers, not
  links,'' 2011. [Online]. Available:
  \url{https://thenextweb.com/google/2011/06/01/google-our-new-search-strategy-is-to-compute-answers-not-links/}
\BIBentrySTDinterwordspacing

\bibitem{Chen2017Reading}
D.~Chen, A.~Fisch, J.~Weston, and A.~Bordes, ``Reading {W}ikipedia to answer
  open-domain questions,'' in \emph{Proceedings of the 55th Annual Meeting of
  the Association for Computational Linguistics (Volume 1: Long Papers)}.\hskip
  1em plus 0.5em minus 0.4em\relax Association for Computational Linguistics,
  2017, pp. 1870--1879.

\bibitem{Voorhees99Trec8}
E.~M. Voorhees, ``The trec-8 question answering track report,'' NIST, Tech.
  Rep., 1999.

\bibitem{Todor2018OpenBookQA}
T.~Mihaylov, P.~Clark, T.~Khot, and A.~Sabharwal, ``Can a suit of armor conduct
  electricity? {A} new dataset for open book question answering,'' \emph{CoRR},
  vol. abs/1809.02789, 2018.

\bibitem{Harabagiu2003Open}
S.~M. Harabagiu, S.~J. Maiorano, and M.~A. Paundefinedca, ``Open-domain textual
  question answering techniques,'' \emph{Nat. Lang. Eng.}, vol.~9, no.~3, p.
  231–267, 2003.

\bibitem{Burger2001Issues}
V.~C. John~Burger, Claire~Cardie \emph{et~al.}, ``Issues, tasks and program
  structures to roadmap research in question \& answering (q \&a,'' NIST, Tech.
  Rep., 2001.

\bibitem{Kolomiyets2011SQASurvey}
O.~Kolomiyets and M.-F. Moens, ``A survey on question answering technology from
  an information retrieval perspective,'' \emph{Inf. Sci.}, vol. 181, no.~24,
  pp. 5412--5434, 2011.

\bibitem{Allam2012QASurvey}
A.~Allam and M.~Haggag, ``The question answering systems: A survey,''
  \emph{International Journal of Research and Reviews in Information Sciences},
  pp. 211--221, 2012.

\bibitem{Mishra2016QASurvey}
A.~Mishra and S.~K. Jain, ``A survey on question answering systems with
  classification,'' \emph{J. King Saud Univ. Comput. Inf. Sci.}, vol.~28,
  no.~3, p. 345–361, 2016.

\bibitem{Pasca2003Open}
M.~Paşca, ``Open-domain question answering from large text collections,''
  \emph{Computational Linguistics}, vol.~29, no.~4, pp. 665--667, 2003.

\bibitem{Huang2020Recent}
Z.~{Huang}, S.~{Xu}, M.~{Hu}, X.~{Wang}, J.~{Qiu}, Y.~{Fu}, Y.~{Zhao},
  Y.~{Peng}, and C.~{Wang}, ``Recent trends in deep learning based open-domain
  textual question answering systems,'' \emph{IEEE Access}, vol.~8, pp.
  94\,341--94\,356, 2020.

\bibitem{Tao2018Novel}
T.~Lei, Z.~Shi, D.~Liu, L.~Yang, and F.~Zhu, ``A novel cnn-based method for
  question classification in intelligent question answering,'' in
  \emph{Proceedings of the 2018 International Conference on Algorithms,
  Computing and Artificial Intelligence}.\hskip 1em plus 0.5em minus
  0.4em\relax Association for Computing Machinery, 2018.

\bibitem{Xia2018Novel}
W.~Xia, W.~Zhu, B.~Liao, M.~Chen, L.~Cai, and L.~Huang, ``Novel architecture
  for long short-term memory used in question classification,''
  \emph{Neurocomputing}, vol. 299, pp. 20--31, 2018.

\bibitem{Nishida2018Retrieve}
K.~Nishida, I.~Saito, A.~Otsuka, H.~Asano, and J.~Tomita, ``Retrieve-and-read:
  Multi-task learning of information retrieval and reading comprehension,'' in
  \emph{Proceedings of the 27th ACM International Conference on Information and
  Knowledge Management}, ser. CIKM ’18.\hskip 1em plus 0.5em minus
  0.4em\relax Association for Computing Machinery, 2018, p. 647–656.

\bibitem{Karpukhin2020Dense}
V.~Karpukhin, B.~O{\u{g}}uz, S.~Min, L.~Wu, S.~Edunov, D.~Chen, and W.-t. Yih,
  ``Dense passage retrieval for open-domain question answering,'' \emph{arXiv
  preprint arXiv:2004.04906}, 2020.

\bibitem{Khattab2020Relevance}
\BIBentryALTinterwordspacing
O.~Khattab, C.~Potts, and M.~Zaharia, ``{Relevance-guided Supervision for
  OpenQA with ColBERT},'' 2020. [Online]. Available:
  \url{http://arxiv.org/abs/2007.00814}
\BIBentrySTDinterwordspacing

\bibitem{Hermann2015Teaching}
K.~M. Hermann, T.~Ko\v{c}isk\'{y}, E.~Grefenstette, L.~Espeholt, W.~Kay,
  M.~Suleyman, and P.~Blunsom, ``Teaching machines to read and comprehend,'' in
  \emph{Proceedings of the 28th International Conference on Neural Information
  Processing Systems - Volume 1}.\hskip 1em plus 0.5em minus 0.4em\relax MIT
  Press, 2015, pp. 1693--1701.

\bibitem{Rajpurkar2016SQuAD}
P.~Rajpurkar, J.~Zhang, K.~Lopyrev, and P.~Liang, ``{SQ}u{AD}: 100,000+
  questions for machine comprehension of text,'' in \emph{Proceedings of the
  2016 Conference on Empirical Methods in Natural Language Processing}.\hskip
  1em plus 0.5em minus 0.4em\relax Association for Computational Linguistics,
  2016, pp. 2383--2392.

\bibitem{Nguyen2016MS}
T.~Nguyen, M.~Rosenberg, X.~Song, J.~Gao, S.~Tiwary, R.~Majumder, and L.~Deng,
  ``{MS} {MARCO:} {A} human generated machine reading comprehension dataset,''
  2016.

\bibitem{Lai2017RACE}
G.~Lai, Q.~Xie, H.~Liu, Y.~Yang, and E.~H. Hovy, ``{RACE:} large-scale reading
  comprehension dataset from examinations,'' \emph{CoRR}, vol. abs/1704.04683,
  2017.

\bibitem{Rajpurkar2018Know}
P.~Rajpurkar, R.~Jia, and P.~Liang, ``Know what you don{'}t know: Unanswerable
  questions for {SQ}u{AD},'' in \emph{Proceedings of the 56th Annual Meeting of
  the Association for Computational Linguistics (Volume 2: Short
  Papers)}.\hskip 1em plus 0.5em minus 0.4em\relax Association for
  Computational Linguistics, 2018, pp. 784--789.

\bibitem{li2020molweni}
J.~Li, M.~Liu, M.-Y. Kan, Z.~Zheng, Z.~Wang, W.~Lei, T.~Liu, and B.~Qin,
  ``Molweni: A challenge multiparty dialogues-based machine reading
  comprehension dataset with discourse structure,'' 2020.

\bibitem{Seo2017Bidirectional}
M.~J. Seo, A.~Kembhavi, A.~Farhadi, and H.~Hajishirzi, ``Bidirectional
  attention flow for machine comprehension,'' in \emph{5th International
  Conference on Learning Representations, {ICLR} 2017}.\hskip 1em plus 0.5em
  minus 0.4em\relax OpenReview.net, 2017.

\bibitem{Wang2017Gated}
W.~Wang, N.~Yang, F.~Wei, B.~Chang, and M.~Zhou, ``Gated self-matching networks
  for reading comprehension and question answering,'' in \emph{Proceedings of
  the 55th Annual Meeting of the Association for Computational Linguistics,
  {ACL}}.\hskip 1em plus 0.5em minus 0.4em\relax Association for Computational
  Linguistics, 2017, pp. 189--198.

\bibitem{Yu2018QANET}
A.~W. Yu, D.~Dohan, M.~Luong, R.~Zhao, K.~Chen, M.~Norouzi, and Q.~V. Le,
  ``Qanet: Combining local convolution with global self-attention for reading
  comprehension,'' in \emph{International Conference on Learning
  Representations, {ICLR}}.\hskip 1em plus 0.5em minus 0.4em\relax
  OpenReview.net, 2018.

\bibitem{Devlin2018Bert}
J.~Devlin, M.~Chang, K.~Lee, and K.~Toutanova, ``{BERT:} pre-training of deep
  bidirectional transformers for language understanding,'' \emph{CoRR}, vol.
  abs/1810.04805, 2018.

\bibitem{Wang2018R3RR}
S.~Wang, M.~Yu, X.~Guo, Z.~Wang, T.~Klinger, W.~Zhang, S.~Chang, G.~Tesauro,
  B.~Zhou, and J.~Jiang, ``R3: Reinforced ranker-reader for open-domain
  question answering,'' in \emph{AAAI}, 2018.

\bibitem{Das2019Multistep}
R.~Das, S.~Dhuliawala, M.~Zaheer, and A.~McCallum, ``Multi-step
  retriever-reader interaction for scalable open-domain question answering,''
  in \emph{International Conference on Learning Representations}, 2019.

\bibitem{Guu2020Realm}
K.~Guu, K.~Lee, Z.~Tung, P.~Pasupat, and M.-W. Chang, ``Realm:
  Retrieval-augmented language model pre-training,'' \emph{CoRR}, 2020.

\bibitem{Ding2019Cognitive}
M.~Ding, C.~Zhou, Q.~Chen, H.~Yang, and J.~Tang, ``Cognitive graph for
  multi-hop reading comprehension at scale,'' in \emph{Proceedings of the 57th
  Annual Meeting of the Association for Computational Linguistics}.\hskip 1em
  plus 0.5em minus 0.4em\relax Association for Computational Linguistics, 2019,
  pp. 2694--2703.

\bibitem{Nie2019Revealing}
Y.~Nie, S.~Wang, and M.~Bansal, ``Revealing the importance of semantic
  retrieval for machine reading at scale,'' in \emph{Proceedings of the 2019
  Conference on Empirical Methods in Natural Language Processing and the 9th
  International Joint Conference on Natural Language Processing
  (EMNLP-IJCNLP)}.\hskip 1em plus 0.5em minus 0.4em\relax Association for
  Computational Linguistics, 2019, pp. 2553--2566.

\bibitem{Lin2018Denoising}
Y.~Lin, H.~Ji, Z.~Liu, and M.~Sun, ``Denoising distantly supervised open-domain
  question answering,'' in \emph{Proceedings of the 56th Annual Meeting of the
  Association for Computational Linguistics (Volume 1: Long Papers)}.\hskip 1em
  plus 0.5em minus 0.4em\relax Association for Computational Linguistics, 2018,
  pp. 1736--1745.

\bibitem{Lee2018Ranking}
J.~Lee, S.~Yun, H.~Kim, M.~Ko, and J.~Kang, ``Ranking paragraphs for improving
  answer recall in open-domain question answering,'' in \emph{Proceedings of
  the 2018 Conference on Empirical Methods in Natural Language
  Processing}.\hskip 1em plus 0.5em minus 0.4em\relax Association for
  Computational Linguistics, 2018, pp. 565--569.

\bibitem{Feldman2019Multihop}
Y.~Feldman \emph{et~al.}, ``Multi-hop paragraph retrieval for open-domain
  question answering,'' in \emph{Proceedings of the 57th Annual Meeting of the
  Association for Computational Linguistics}.\hskip 1em plus 0.5em minus
  0.4em\relax Association for Computational Linguistics, 2019, pp. 2296--2309.

\bibitem{Qi2019Answering}
P.~Qi, X.~Lin, L.~Mehr, Z.~Wang, and C.~D. Manning, ``Answering complex
  open-domain questions through iterative query generation,'' in
  \emph{Proceedings of the 2019 Conference on Empirical Methods in Natural
  Language Processing and the 9th International Joint Conference on Natural
  Language Processing (EMNLP-IJCNLP)}.\hskip 1em plus 0.5em minus 0.4em\relax
  Association for Computational Linguistics, 2019, pp. 2590--2602.

\bibitem{Lee2019Latent}
K.~Lee, M.-W. Chang, and K.~Toutanova, ``Latent retrieval for weakly supervised
  open domain question answering,'' in \emph{Proceedings of the 57th Annual
  Meeting of the Association for Computational Linguistics}.\hskip 1em plus
  0.5em minus 0.4em\relax Association for Computational Linguistics, 2019, pp.
  6086--6096.

\bibitem{Lewis2020Retrieval}
\BIBentryALTinterwordspacing
P.~Lewis, E.~Perez, A.~Piktus, F.~Petroni, V.~Karpukhin, N.~Goyal,
  H.~K{\"{u}}ttler, M.~Lewis, W.-t. Yih, T.~Rockt{\"{a}}schel, S.~Riedel, and
  D.~Kiela, ``{Retrieval-Augmented Generation for Knowledge-Intensive NLP
  Tasks},'' 2020. [Online]. Available: \url{http://arxiv.org/abs/2005.11401}
\BIBentrySTDinterwordspacing

\bibitem{Woods1973LUNAR}
W.~A. Woods, ``Progress in natural language understanding: An application to
  lunar geology,'' in \emph{Proceedings of the June 4-8, 1973, National
  Computer Conference and Exposition}.\hskip 1em plus 0.5em minus 0.4em\relax
  ACM, 1973, pp. 441--450.

\bibitem{Kupiec1993MURAX}
J.~Kupiec, ``Murax: A robust linguistic approach for question answering using
  an on-line encyclopedia,'' in \emph{Proceedings of the 16th Annual
  International ACM SIGIR Conference on Research and Development in Information
  Retrieval}.\hskip 1em plus 0.5em minus 0.4em\relax Association for Computing
  Machinery, 1993, p. 181–190.

\bibitem{Voorhees2001Trec}
E.~M. Voorhees, ``Overview of the trec 2001 question answering track,'' in
  \emph{In Proceedings of TREC-10}, 2001, pp. 42--51.

\bibitem{Voorhees2003Trec11}
------, ``Overview of the {TREC} 2002 question answering track,'' in
  \emph{Proceedings of The Eleventh Text REtrieval Conference, {TREC} 2002,
  Gaithersburg, Maryland, USA, November 19-22, 2002}, ser. {NIST} Special
  Publication, vol. 500-251.\hskip 1em plus 0.5em minus 0.4em\relax National
  Institute of Standards and Technology {(NIST)}, 2002.

\bibitem{Voorhees2003Trec12}
E.~Voorhees, ``Overview of the trec 2003 question answering track,'' NIST,
  Tech. Rep., 2003.

\bibitem{Kwok2001Scaling}
C.~Kwok, O.~Etzioni, O.~Etzioni, and D.~S. Weld, ``Scaling question answering
  to the web,'' \emph{ACM Transactions on Information Systems}, vol.~19, no.~3,
  pp. 242--262, 2001.

\bibitem{Brill2002AskMSR}
E.~Brill, S.~Dumais, and M.~Banko, ``An analysis of the {A}sk{MSR}
  question-answering system,'' in \emph{Proceedings of the 2002 Conference on
  Empirical Methods in Natural Language Processing ({EMNLP} 2002)}.\hskip 1em
  plus 0.5em minus 0.4em\relax Association for Computational Linguistics, 2002,
  pp. 257--264.

\bibitem{Zheng2002AnswerBusQA}
Z.~Zheng, ``Answerbus question answering system,'' in \emph{Proceedings of the
  Second International Conference on Human Language Technology Research}, ser.
  HLT '02.\hskip 1em plus 0.5em minus 0.4em\relax Morgan Kaufmann Publishers
  Inc., 2002, p. 399–404.

\bibitem{Moldovan2002Performance}
D.~Moldovan, M.~Pasca, S.~Harabagiu, and M.~Surdeanu, ``Performance issues and
  error analysis in an open-domain question answering system,'' in
  \emph{Proceedings of the 40th Annual Meeting of the Association for
  Computational Linguistics}.\hskip 1em plus 0.5em minus 0.4em\relax
  Association for Computational Linguistics, 2002, pp. 33--40.

\bibitem{Sun2005Using}
R.~Sun, J.~Jiang, Y.~F. Tan, H.~Cui, T.-S. Chua, and M.-Y. Kan, ``Using
  syntactic and semantic relation analysis in question answering,'' in
  \emph{TREC}, 2005.

\bibitem{Xu1996QELG}
J.~Xu and W.~B. Croft, ``Query expansion using local and global document
  analysis,'' in \emph{Proceedings of the 19th Annual International ACM SIGIR
  Conference on Research and Development in Information Retrieval}.\hskip 1em
  plus 0.5em minus 0.4em\relax Association for Computing Machinery, 1996, p.
  4–11.

\bibitem{Carpineto2012SurveyAQE}
C.~Carpineto and G.~Romano, ``A survey of automatic query expansion in
  information retrieval,'' \emph{ACM Computing Survey}, vol.~44, no.~1, 2012.

\bibitem{Quirk2004Monolingual}
C.~Quirk, C.~Brockett, and W.~Dolan, ``Monolingual machine translation for
  paraphrase generation,'' in \emph{Proceedings of the 2004 Conference on
  Empirical Methods in Natural Language Processing}.\hskip 1em plus 0.5em minus
  0.4em\relax Association for Computational Linguistics, 2004, pp. 142--149.

\bibitem{Bannard2005Paraphrasing}
C.~Bannard and C.~Callison-Burch, ``Paraphrasing with bilingual parallel
  corpora,'' in \emph{Proceedings of the 43rd Annual Meeting of the Association
  for Computational Linguistics ({ACL}{'}05)}.\hskip 1em plus 0.5em minus
  0.4em\relax Association for Computational Linguistics, 2005, pp. 597--604.

\bibitem{Zhao2008Combining}
S.~Zhao, C.~Niu, M.~Zhou, T.~Liu, and S.~Li, ``Combining multiple resources to
  improve {SMT}-based paraphrasing model,'' in \emph{Proceedings of ACL-08:
  HLT}.\hskip 1em plus 0.5em minus 0.4em\relax Association for Computational
  Linguistics, 2008, pp. 1021--1029.

\bibitem{Wubben2010Paraphrase}
S.~Wubben, A.~van~den Bosch, and E.~Krahmer, ``Paraphrase generation as
  monolingual translation: Data and evaluation,'' in \emph{Proceedings of the
  6th International Natural Language Generation Conference}, 2010.

\bibitem{Li2002Learning}
X.~Li and D.~Roth, ``Learning question classifiers,'' in \emph{{COLING} 2002:
  The 19th International Conference on Computational Linguistics}, 2002.

\bibitem{Suzuki2003Question}
J.~Suzuki, H.~Taira, Y.~Sasaki, and E.~Maeda, ``Question classification using
  {HDAG} kernel,'' in \emph{Proceedings of the {ACL} 2003 Workshop on
  Multilingual Summarization and Question Answering}.\hskip 1em plus 0.5em
  minus 0.4em\relax Association for Computational Linguistics, 2003, pp.
  61--68.

\bibitem{Zhang2003Question}
D.~Zhang and W.~S. Lee, ``Question classification using support vector
  machines,'' in \emph{Proceedings of the 26th Annual International ACM SIGIR
  Conference on Research and Development in Informaion Retrieval}, ser. SIGIR
  ’03.\hskip 1em plus 0.5em minus 0.4em\relax Association for Computing
  Machinery, 2003, p. 26–32.

\bibitem{Ferrucci2010Watson}
D.~Ferrucci, E.~Brown, J.~Chu-Carroll, J.~Fan, D.~Gondek, A.~A. Kalyanpur,
  A.~Lally, J.~W. Murdock, E.~Nyberg, J.~Prager, N.~Schlaefer, and C.~Welty,
  ``Building watson: An overview of the deepqa project,'' \emph{AI Magazine},
  vol.~31, no.~3, pp. 59--79, 2010.

\bibitem{Madabushi2016High}
H.~Tayyar~Madabushi and M.~Lee, ``High accuracy rule-based question
  classification using question syntax and semantics,'' in \emph{Proceedings of
  {COLING} 2016, the 26th International Conference on Computational
  Linguistics: Technical Papers}.\hskip 1em plus 0.5em minus 0.4em\relax The
  COLING 2016 Organizing Committee, 2016, pp. 1220--1230.

\bibitem{Manning2008IR}
C.~D. Manning, P.~Raghavan, and H.~Sch\"{u}tze, \emph{Introduction to
  Information Retrieval}.\hskip 1em plus 0.5em minus 0.4em\relax USA: Cambridge
  University Press, 2008.

\bibitem{Robertson2009probabilistic}
S.~Robertson, H.~Zaragoza \emph{et~al.}, ``The probabilistic relevance
  framework: Bm25 and beyond,'' \emph{Foundations and Trends in Information
  Retrieval}, vol.~3, no.~4, pp. 333--389, 2009.

\bibitem{Kluwer2003Language}
W.~B. Croft and J.~Lafferty, \emph{Language modeling for information
  retrieval}.\hskip 1em plus 0.5em minus 0.4em\relax Kluwer Academic Publ.,
  2003.

\bibitem{Voorhees2004Trec}
E.~M. Voorhees, ``Overview of the trec 2004 question answering track,'' in
  \emph{In Proceedings of the Thirteenth Text REtreival Conference (TREC
  2004)}, 2005, pp. 52--62.

\bibitem{Molla2006Named}
D.~Moll{\'a}, M.~van Zaanen, and D.~Smith, ``Named entity recognition for
  question answering,'' in \emph{Proceedings of the Australasian Language
  Technology Workshop 2006}, 2006, pp. 51--58.

\bibitem{Wang2006ASO}
M.~Wang, ``A survey of answer extraction techniques in factoid question
  answering,'' \emph{Computational Linguistics}, vol.~1, no.~1, pp. 1--14,
  2006.

\bibitem{Soubbotin2001Patterns}
M.~M. Soubbotin and S.~M. Soubbotin, ``Patterns of potential answer expressions
  as clues to the right answers,'' in \emph{In Proceedings of the 10th Text
  REtrieval Conference (TREC-10)}, 2001.

\bibitem{Ravichandran2002Learning}
D.~Ravichandran and E.~Hovy, ``Learning surface text patterns for a question
  answering system,'' in \emph{Proceedings of the 40th Annual Meeting of the
  Association for Computational Linguistics}.\hskip 1em plus 0.5em minus
  0.4em\relax Association for Computational Linguistics, 2002, pp. 41--47.

\bibitem{Shen2005Exploring}
D.~Shen, G.-J.~M. Kruijff, and D.~Klakow, ``Exploring syntactic relation
  patterns for question answering,'' in \emph{Second International Joint
  Conference on Natural Language Processing: Full Papers}, 2005.

\bibitem{Ram2012Maximum}
P.~Ram and A.~G. Gray, ``Maximum inner-product search using cone trees,'' in
  \emph{Proceedings of the 18th ACM SIGKDD international conference on
  Knowledge discovery and data mining}, 2012, pp. 931--939.

\bibitem{Shrivastava2014ALSH}
A.~Shrivastava and P.~Li, ``Asymmetric lsh (alsh) for sublinear time maximum
  inner product search (mips),'' in \emph{Advances in Neural Information
  Processing Systems}, 2014, pp. 2321--2329.

\bibitem{Shen2015Learning}
F.~Shen, W.~Liu, S.~Zhang, Y.~Yang, and H.~Tao~Shen, ``Learning binary codes
  for maximum inner product search,'' in \emph{Proceedings of the IEEE
  International Conference on Computer Vision}, 2015, pp. 4148--4156.

\bibitem{Vaswani2017Attention}
A.~Vaswani, N.~Shazeer, N.~Parmar, J.~Uszkoreit, L.~Jones, A.~N. Gomez,
  L.~Kaiser, and I.~Polosukhin, ``Attention is all you need,'' in
  \emph{Advances in Neural Information Processing Systems 30: Annual Conference
  on Neural Information Processing Systems 2017, 4-9 December 2017, Long Beach,
  CA, {USA}}, 2017, pp. 5998--6008.

\bibitem{Seo2019Real}
M.~Seo, J.~Lee, T.~Kwiatkowski, A.~Parikh, A.~Farhadi, and H.~Hajishirzi,
  ``Real-time open-domain question answering with dense-sparse phrase index,''
  in \emph{Proceedings of the 57th Annual Meeting of the Association for
  Computational Linguistics}.\hskip 1em plus 0.5em minus 0.4em\relax
  Association for Computational Linguistics, 2019, pp. 4430--4441.

\bibitem{Dehghani2019Learning}
M.~Dehghani, H.~Azarbonyad, J.~Kamps, and M.~de~Rijke, ``Learning to transform,
  combine, and reason in open-domain question answering,'' in \emph{Proceedings
  of the Twelfth ACM International Conference on Web Search and Data Mining},
  ser. WSDM ’19.\hskip 1em plus 0.5em minus 0.4em\relax Association for
  Computing Machinery, 2019, p. 681–689.

\bibitem{Dhingra2020Differentiable}
B.~Dhingra, M.~Zaheer, V.~Balachandran, G.~Neubig, R.~Salakhutdinov, and W.~W.
  Cohen, ``Differentiable reasoning over a virtual knowledge base,'' in
  \emph{International Conference on Learning Representations}, 2020.

\bibitem{Kratzwald2019RankQA}
B.~Kratzwald, A.~Eigenmann, and S.~Feuerriegel, ``{R}ank{QA}: Neural question
  answering with answer re-ranking,'' in \emph{Proceedings of the 57th Annual
  Meeting of the Association for Computational Linguistics}.\hskip 1em plus
  0.5em minus 0.4em\relax Association for Computational Linguistics, 2019, pp.
  6076--6085.

\bibitem{Kratzwald2018Adaptive}
B.~Kratzwald \emph{et~al.}, ``Adaptive document retrieval for deep question
  answering,'' in \emph{Proceedings of the 2018 Conference on Empirical Methods
  in Natural Language Processing}.\hskip 1em plus 0.5em minus 0.4em\relax
  Association for Computational Linguistics, 2018, pp. 576--581.

\bibitem{Yang2019End2end}
W.~Yang, Y.~Xie, A.~Lin, X.~Li, L.~Tan, K.~Xiong, M.~Li, and J.~Lin,
  ``End-to-end open-domain question answering with {BERT}serini,'' in
  \emph{Proceedings of the 2019 Conference of the North {A}merican Chapter of
  the Association for Computational Linguistics (Demonstrations)}.\hskip 1em
  plus 0.5em minus 0.4em\relax Association for Computational Linguistics, 2019,
  pp. 72--77.

\bibitem{Wang2019Multipassage}
Z.~Wang, P.~Ng, X.~Ma, R.~Nallapati, and B.~Xiang, ``Multi-passage {BERT}: A
  globally normalized {BERT} model for open-domain question answering,'' in
  \emph{Proceedings of the 2019 Conference on Empirical Methods in Natural
  Language Processing and the 9th International Joint Conference on Natural
  Language Processing (EMNLP-IJCNLP)}.\hskip 1em plus 0.5em minus 0.4em\relax
  Association for Computational Linguistics, 2019, pp. 5878--5882.

\bibitem{Weinberger2009Feature}
K.~Weinberger, A.~Dasgupta, J.~Langford, A.~Smola, and J.~Attenberg, ``Feature
  hashing for large scale multitask learning,'' in \emph{Proceedings of the
  26th Annual International Conference on Machine Learning}.\hskip 1em plus
  0.5em minus 0.4em\relax Association for Computing Machinery, 2009, p.
  1113–1120.

\bibitem{Yang2017Anserini}
P.~Yang, H.~Fang, and J.~Lin, ``Anserini: Enabling the use of lucene for
  information retrieval research,'' in \emph{Proceedings of the 40th
  International ACM SIGIR Conference on Research and Development in Information
  Retrieval}, ser. SIGIR ’17.\hskip 1em plus 0.5em minus 0.4em\relax
  Association for Computing Machinery, 2017, p. 1253–1256.

\bibitem{Zhao2020SPARTA}
T.~Zhao, X.~Lu, and K.~Lee, ``Sparta: Efficient open-domain question answering
  via sparse transformer matching retrieval,'' \emph{arXiv preprint
  arXiv:2009.13013}, 2020.

\bibitem{zhang2020dcbert}
Y.~Zhang, P.~Nie, X.~Geng, A.~Ramamurthy, L.~Song, and D.~Jiang, ``Dc-bert:
  Decoupling question and document for efficient contextual encoding,'' 2020.

\bibitem{Khattab2020ColBERT}
O.~Khattab and M.~Zaharia, ``Colbert: Efficient and effective passage search
  via contextualized late interaction over bert,'' in \emph{Proceedings of the
  43rd International ACM SIGIR Conference on Research and Development in
  Information Retrieval}, ser. SIGIR '20.\hskip 1em plus 0.5em minus
  0.4em\relax Association for Computing Machinery, 2020, p. 39–48.

\bibitem{Xiong2020Answering}
W.~Xiong, X.~L. Li, S.~Iyer, J.~Du, P.~Lewis, W.~Y. Wang, Y.~Mehdad, W.-t. Yih,
  S.~Riedel, D.~Kiela \emph{et~al.}, ``Answering complex open-domain questions
  with multi-hop dense retrieval,'' \emph{arXiv preprint arXiv:2009.12756},
  2020.

\bibitem{Zhang2020DDRQA}
Y.~Zhang, P.~Nie, A.~Ramamurthy, and L.~Song, ``Ddrqa: Dynamic document
  reranking for open-domain multi-hop question answering,'' \emph{arXiv
  preprint arXiv:2009.07465}, 2020.

\bibitem{Mao2020Generation}
\BIBentryALTinterwordspacing
Y.~Mao, P.~He, X.~Liu, Y.~Shen, J.~Gao, J.~Han, and W.~Chen,
  ``{Generation-Augmented Retrieval for Open-domain Question Answering},''
  2020. [Online]. Available: \url{http://arxiv.org/abs/2009.08553}
\BIBentrySTDinterwordspacing

\bibitem{Asai2020Learning}
A.~Asai, K.~Hashimoto, H.~Hajishirzi, R.~Socher, and C.~Xiong, ``Learning to
  retrieve reasoning paths over wikipedia graph for question answering,'' in
  \emph{International Conference on Learning Representations}, 2020.

\bibitem{Min2019Knowledge}
\BIBentryALTinterwordspacing
S.~Min, D.~Chen, L.~Zettlemoyer, and H.~Hajishirzi, ``{Knowledge Guided Text
  Retrieval and Reading for Open Domain Question Answering},'' 2019. [Online].
  Available: \url{http://arxiv.org/abs/1911.03868}
\BIBentrySTDinterwordspacing

\bibitem{Yang2018Hotpotqa}
Z.~Yang, P.~Qi, S.~Zhang, Y.~Bengio, W.~Cohen, R.~Salakhutdinov, and C.~D.
  Manning, ``{H}otpot{QA}: A dataset for diverse, explainable multi-hop
  question answering,'' in \emph{Proceedings of the 2018 Conference on
  Empirical Methods in Natural Language Processing}.\hskip 1em plus 0.5em minus
  0.4em\relax Association for Computational Linguistics, 2018, pp. 2369--2380.

\bibitem{Welbl2018Constructing}
\BIBentryALTinterwordspacing
J.~Welbl, P.~Stenetorp, and S.~Riedel, ``Constructing datasets for multi-hop
  reading comprehension across documents,'' \emph{Transactions of the
  Association for Computational Linguistics}, pp. 287--302, 2018. [Online].
  Available: \url{https://www.aclweb.org/anthology/Q18-1021}
\BIBentrySTDinterwordspacing

\bibitem{Lewis2020Bart}
M.~Lewis, Y.~Liu, N.~Goyal, M.~Ghazvininejad, A.~Mohamed, O.~Levy, V.~Stoyanov,
  and L.~Zettlemoyer, ``{BART}: Denoising sequence-to-sequence pre-training for
  natural language generation, translation, and comprehension,'' in
  \emph{Proceedings of the 58th Annual Meeting of the Association for
  Computational Linguistics}.\hskip 1em plus 0.5em minus 0.4em\relax
  Association for Computational Linguistics, 2020, pp. 7871--7880.

\bibitem{Cho2014Learning}
K.~Cho, B.~van Merrienboer, {\c{C}}.~G{\"{u}}l{\c{c}}ehre, D.~Bahdanau,
  F.~Bougares, H.~Schwenk, and Y.~Bengio, ``Learning phrase representations
  using {RNN} encoder-decoder for statistical machine translation,'' in
  \emph{EMNLP}.\hskip 1em plus 0.5em minus 0.4em\relax {ACL}, 2014, pp.
  1724--1734.

\bibitem{Clark2018Simple}
C.~Clark and M.~Gardner, ``Simple and effective multi-paragraph reading
  comprehension,'' in \emph{Proceedings of the 56th Annual Meeting of the
  Association for Computational Linguistics, {ACL}}.\hskip 1em plus 0.5em minus
  0.4em\relax Association for Computational Linguistics, 2018, pp. 845--855.

\bibitem{Andrew2004Quick}
A.~Lampert, ``A quick introduction to question answering,'' \emph{Dated
  December}, 2004.

\bibitem{Htut2018Training}
P.~M. Htut, S.~Bowman, and K.~Cho, ``Training a ranking function for
  open-domain question answering,'' in \emph{Proceedings of the 2018 Conference
  of the North {A}merican Chapter of the Association for Computational
  Linguistics: Student Research Workshop}.\hskip 1em plus 0.5em minus
  0.4em\relax Association for Computational Linguistics, 2018, pp. 120--127.

\bibitem{Banerjee2019Careful}
P.~Banerjee, K.~K. Pal, A.~Mitra, and C.~Baral, ``Careful selection of
  knowledge to solve open book question answering,'' in \emph{Proceedings of
  the 57th Annual Meeting of the Association for Computational
  Linguistics}.\hskip 1em plus 0.5em minus 0.4em\relax Association for
  Computational Linguistics, 2019, pp. 6120--6129.

\bibitem{Wang2020Answering}
J.~Wang, A.~Jatowt, M.~F{\"{a}}rber, and M.~Yoshikawa, ``Answering
  event-related questions over long-term news article archives,'' in
  \emph{ECIR}, ser. Lecture Notes in Computer Science, vol. 12035.\hskip 1em
  plus 0.5em minus 0.4em\relax Springer, 2020, pp. 774--789.

\bibitem{Wang2021Improving}
J.~Wang, A.~Jatowt, M.~F{\"a}rber, and M.~Yoshikawa, ``{Improving question
  answering for event-focused questions in temporal collections of news
  articles},'' \emph{Information Retrieval Journal}, vol.~24, no.~1, pp.
  29--54, 2021.

\bibitem{Conneau2017Supervised}
A.~Conneau, D.~Kiela, H.~Schwenk, L.~Barrault, and A.~Bordes, ``Supervised
  learning of universal sentence representations from natural language
  inference data,'' in \emph{Proceedings of the 2017 Conference on Empirical
  Methods in Natural Language Processing}.\hskip 1em plus 0.5em minus
  0.4em\relax Association for Computational Linguistics, 2017, pp. 670--680.

\bibitem{Santoro2017Simple}
A.~Santoro, D.~Raposo, D.~G. Barrett, M.~Malinowski, R.~Pascanu, P.~Battaglia,
  and T.~Lillicrap, ``A simple neural network module for relational
  reasoning,'' in \emph{Advances in Neural Information Processing Systems 30},
  I.~Guyon, U.~V. Luxburg, S.~Bengio, H.~Wallach, R.~Fergus, S.~Vishwanathan,
  and R.~Garnett, Eds.\hskip 1em plus 0.5em minus 0.4em\relax Curran
  Associates, Inc., 2017, pp. 4967--4976.

\bibitem{Richardson2013MCTest}
M.~Richardson, C.~J. Burges, and E.~Renshaw, ``{MCT}est: A challenge dataset
  for the open-domain machine comprehension of text,'' in \emph{Proceedings of
  the 2013 Conference on Empirical Methods in Natural Language
  Processing}.\hskip 1em plus 0.5em minus 0.4em\relax Association for
  Computational Linguistics, 2013, pp. 193--203.

\bibitem{Dua2019DROP}
D.~Dua, Y.~Wang, P.~Dasigi, G.~Stanovsky, S.~Singh, and M.~Gardner, ``{DROP}: A
  reading comprehension benchmark requiring discrete reasoning over
  paragraphs,'' in \emph{Proc. of NAACL}, 2019.

\bibitem{Izacard2020Leveraging}
G.~Izacard and E.~Grave, ``Leveraging passage retrieval with generative models
  for open domain question answering,'' \emph{arXiv preprint arXiv:2007.01282},
  2020.

\bibitem{Kipf2017Semi}
T.~N. Kipf and M.~Welling, ``Semi-supervised classification with graph
  convolutional networks,'' in \emph{ICLR}, 2017.

\bibitem{Joshi2019Spanbert}
M.~Joshi, D.~Chen, Y.~Liu, D.~S. Weld, L.~Zettlemoyer, and O.~Levy,
  ``{SpanBERT}: Improving pre-training by representing and predicting spans,''
  \emph{arXiv preprint arXiv:1907.10529}, 2019.

\bibitem{Tang2018S-Net}
C.~Tan, F.~Wei, N.~Yang, B.~Du, W.~Lv, and M.~Zhou, ``S-net: From answer
  extraction to answer synthesis for machine reading comprehension,'' in
  \emph{AAAI}.\hskip 1em plus 0.5em minus 0.4em\relax {AAAI} Press, 2018, pp.
  5940--5947.

\bibitem{Raffel2019T5}
C.~Raffel, N.~Shazeer, A.~Roberts, K.~Lee, S.~Narang, M.~Matena, Y.~Zhou,
  W.~Li, and P.~J. Liu, ``Exploring the limits of transfer learning with a
  unified text-to-text transformer,'' \emph{arXiv e-prints}, 2019.

\bibitem{Liu2019Neural}
S.~Liu, X.~Zhang, S.~Zhang, H.~Wang, and W.~Zhang, ``Neural machine reading
  comprehension: Methods and trends,'' \emph{CoRR}, vol. abs/1907.01118, 2019.

\bibitem{Wang2018Evidence}
S.~Wang, M.~Yu, J.~Jiang, W.~Zhang, X.~Guo, S.~Chang, Z.~Wang, T.~Klinger,
  G.~Tesauro, and M.~Campbell, ``Evidence aggregation for answer re-ranking in
  open-domain question answering,'' in \emph{6th International Conference on
  Learning Representations, {ICLR} 2018, Vancouver, BC, Canada, April 30 - May
  3, 2018, Conference Track Proceedings}.\hskip 1em plus 0.5em minus
  0.4em\relax ICLR, 2018.

\bibitem{Wang2016Learning}
S.~Wang and J.~Jiang, ``Learning natural language inference with {LSTM},'' in
  \emph{Conference of the North American Chapter of the Association for
  Computational Linguistics: Human Language Technologies}.\hskip 1em plus 0.5em
  minus 0.4em\relax The Association for Computational Linguistics, 2016, pp.
  1442--1451.

\bibitem{radford2019GPT2}
A.~Radford, J.~Wu, R.~Child, D.~Luan, D.~Amodei, and I.~Sutskever, ``Language
  models are unsupervised multitask learners,'' \emph{OpenAI blog}, vol.~1,
  no.~8, p.~9, 2019.

\bibitem{Brown2020GPT3}
T.~B. Brown, B.~Mann, N.~Ryder, M.~Subbiah, J.~Kaplan, P.~Dhariwal,
  A.~Neelakantan, P.~Shyam, G.~Sastry, A.~Askell \emph{et~al.}, ``Language
  models are few-shot learners,'' \emph{arXiv preprint arXiv:2005.14165}, 2020.

\bibitem{Jeff2017Billion}
J.~Johnson, M.~Douze, and H.~J{\'{e}}gou, ``Billion-scale similarity search
  with gpus,'' \emph{CoRR}, vol. abs/1702.08734, 2017.

\bibitem{Petroni2020Language}
F.~Petroni, T.~Rockt{\"a}schel, P.~Lewis, A.~Bakhtin, Y.~Wu, A.~H. Miller, and
  S.~Riedel, ``Language models as knowledge bases?'' \emph{arXiv preprint
  arXiv:1909.01066}, 2019.

\bibitem{Roberts2020How}
A.~Roberts, C.~Raffel, and N.~Shazeer, ``How much knowledge can you pack into
  the parameters of a language model?'' \emph{arXiv preprint arXiv:2002.08910},
  2020.

\bibitem{Seo2018Phrase}
M.~Seo, T.~Kwiatkowski, A.~Parikh, A.~Farhadi, and H.~Hajishirzi,
  ``Phrase-indexed question answering: A new challenge for scalable document
  comprehension,'' in \emph{Proceedings of the 2018 Conference on Empirical
  Methods in Natural Language Processing}.\hskip 1em plus 0.5em minus
  0.4em\relax Association for Computational Linguistics, 2018, pp. 559--564.

\bibitem{Auer2007DBpedia}
S.~Auer, C.~Bizer, G.~Kobilarov, J.~Lehmann, R.~Cyganiak, and Z.~Ives,
  ``Dbpedia: A nucleus for a web of open data,'' in \emph{The Semantic
  Web}.\hskip 1em plus 0.5em minus 0.4em\relax Springer Berlin Heidelberg,
  2007, pp. 722--735.

\bibitem{Bollacker2008Freebase}
K.~Bollacker, C.~Evans, P.~Paritosh, T.~Sturge, and J.~Taylor, ``Freebase: A
  collaboratively created graph database for structuring human knowledge,'' in
  \emph{Proceedings of the 2008 ACM SIGMOD International Conference on
  Management of Data}.\hskip 1em plus 0.5em minus 0.4em\relax ACM, 2008, pp.
  1247--1250.

\bibitem{Hoffart2013YAGO2}
J.~Hoffart, F.~M. Suchanek, K.~Berberich, and G.~Weikum, ``Yago2: A spatially
  and temporally enhanced knowledge base from wikipedia,'' \emph{Artif.
  Intell.}, vol. 194, pp. 28--61, 2013.

\bibitem{Sun2015Open}
H.~Sun, H.~Ma, W.-t. Yih, C.-T. Tsai, J.~Liu, and M.-W. Chang, ``Open domain
  question answering via semantic enrichment,'' in \emph{Proceedings of the
  24th International Conference on World Wide Web}.\hskip 1em plus 0.5em minus
  0.4em\relax International World Wide Web Conferences Steering Committee,
  2015, pp. 1045--1055.

\bibitem{Sun2018Open}
H.~Sun, B.~Dhingra, M.~Zaheer, K.~Mazaitis, R.~Salakhutdinov, and W.~Cohen,
  ``Open domain question answering using early fusion of knowledge bases and
  text,'' in \emph{Proceedings of the 2018 Conference on Empirical Methods in
  Natural Language Processing}.\hskip 1em plus 0.5em minus 0.4em\relax
  Association for Computational Linguistics, 2018, pp. 4231--4242.

\bibitem{Sun2019Pullnet}
H.~Sun, T.~Bedrax-Weiss, and W.~Cohen, ``{P}ull{N}et: Open domain question
  answering with iterative retrieval on knowledge bases and text,'' in
  \emph{Proceedings of the 2019 Conference on Empirical Methods in Natural
  Language Processing and the 9th International Joint Conference on Natural
  Language Processing (EMNLP-IJCNLP)}.\hskip 1em plus 0.5em minus 0.4em\relax
  Association for Computational Linguistics, 2019, pp. 2380--2390.

\bibitem{Li2016Gated}
Y.~Li, D.~Tarlow, M.~Brockschmidt, and R.~S. Zemel, ``Gated graph sequence
  neural networks,'' in \emph{ICLR}, 2016.

\bibitem{Scarselli2009GNN}
F.~{Scarselli}, M.~{Gori}, A.~C. {Tsoi}, M.~{Hagenbuchner}, and
  G.~{Monfardini}, ``The graph neural network model,'' \emph{IEEE Transactions
  on Neural Networks}, vol.~20, no.~1, pp. 61--80, 2009.

\bibitem{Dai2019XLNet}
Z.~Yang, Z.~Dai, Y.~Yang, J.~G. Carbonell, R.~Salakhutdinov, and Q.~V. Le,
  ``Xlnet: Generalized autoregressive pretraining for language understanding,''
  \emph{CoRR}, vol. abs/1906.08237, 2019.

\bibitem{Hill2015CBT}
F.~Hill, A.~Bordes, S.~Chopra, and J.~Weston, ``The goldilocks principle:
  Reading children's books with explicit memory representations,'' \emph{CoRR},
  2015.

\bibitem{Trischler2017Newsqa}
A.~Trischler, T.~Wang, X.~Yuan, J.~Harris, A.~Sordoni, P.~Bachman, and
  K.~Suleman, ``{N}ews{QA}: A machine comprehension dataset,'' in
  \emph{Proceedings of the 2nd Workshop on Representation Learning for
  {NLP}}.\hskip 1em plus 0.5em minus 0.4em\relax Association for Computational
  Linguistics, 2017, pp. 191--200.

\bibitem{Dunn2017SearchQA}
M.~Dunn, L.~Sagun, M.~Higgins, V.~U. G{\"{u}}ney, V.~Cirik, and K.~Cho,
  ``Searchqa: {A} new q{\&}a dataset augmented with context from a search
  engine,'' \emph{CoRR}, vol. abs/1704.05179, 2017.

\bibitem{Joshi2017Triviaqa}
M.~Joshi, E.~Choi, D.~Weld, and L.~Zettlemoyer, ``{T}rivia{QA}: A large scale
  distantly supervised challenge dataset for reading comprehension,'' in
  \emph{Proceedings of the 55th Annual Meeting of the Association for
  Computational Linguistics (Volume 1: Long Papers)}.\hskip 1em plus 0.5em
  minus 0.4em\relax Association for Computational Linguistics, 2017, pp.
  1601--1611.

\bibitem{Tom2017NarrativeQA}
T.~Kocisk{\'{y}}, J.~Schwarz, P.~Blunsom, C.~Dyer, K.~M. Hermann, G.~Melis, and
  E.~Grefenstette, ``The narrativeqa reading comprehension challenge,''
  \emph{CoRR}, vol. abs/1712.07040, 2017.

\bibitem{He2017DuReader}
W.~He, K.~Liu, Y.~Lyu, S.~Zhao, X.~Xiao, Y.~Liu, Y.~Wang, H.~Wu, Q.~She,
  X.~Liu, T.~Wu, and H.~Wang, ``Dureader: a chinese machine reading
  comprehension dataset from real-world applications,'' \emph{CoRR}, vol.
  abs/1711.05073, 2017.

\bibitem{Reddy2018CoQA}
S.~Reddy, D.~Chen, and C.~D. Manning, ``Coqa: {A} conversational question
  answering challenge,'' \emph{CoRR}, vol. abs/1808.07042, 2018.

\bibitem{Choi2018QuAC}
E.~Choi, H.~He, M.~Iyyer, M.~Yatskar, W.~Yih, Y.~Choi, P.~Liang, and
  L.~Zettlemoyer, ``Quac : Question answering in context,'' \emph{CoRR}, vol.
  abs/1808.07036, 2018.

\bibitem{Clark2018Think}
P.~Clark, I.~Cowhey, O.~Etzioni, T.~Khot, A.~Sabharwal, C.~Schoenick, and
  O.~Tafjord, ``Think you have solved question answering? try arc, the {AI2}
  reasoning challenge,'' \emph{CoRR}, vol. abs/1803.05457, 2018.

\bibitem{Saeidi2018ShARC}
M.~Saeidi, M.~Bartolo, P.~Lewis, S.~Singh, T.~Rockt{\"a}schel, M.~Sheldon,
  G.~Bouchard, and S.~Riedel, ``Interpretation of natural language rules in
  conversational machine reading,'' in \emph{Proceedings of the 2018 Conference
  on Empirical Methods in Natural Language Processing}.\hskip 1em plus 0.5em
  minus 0.4em\relax Association for Computational Linguistics, 2018, pp.
  2087--2097.

\bibitem{Suster2018Clicr}
S.~{\v{S}}uster \emph{et~al.}, ``{C}li{CR}: a dataset of clinical case reports
  for machine reading comprehension,'' in \emph{Proceedings of the 2018
  Conference of the North {A}merican Chapter of the Association for
  Computational Linguistics: Human Language Technologies, Volume 1 (Long
  Papers)}.\hskip 1em plus 0.5em minus 0.4em\relax Association for
  Computational Linguistics, 2018, pp. 1551--1563.

\bibitem{Khashabi2018Looking}
D.~Khashabi, S.~Chaturvedi, M.~Roth, S.~Upadhyay, and D.~Roth, ``Looking beyond
  the surface: A challenge set for reading comprehension over multiple
  sentences,'' in \emph{Proceedings of the 2018 Conference of the North
  {A}merican Chapter of the Association for Computational Linguistics: Human
  Language Technologies, Volume 1 (Long Papers)}.\hskip 1em plus 0.5em minus
  0.4em\relax Association for Computational Linguistics, 2018, pp. 252--262.

\bibitem{Zellers2018SWAG}
R.~Zellers, Y.~Bisk, R.~Schwartz, and Y.~Choi, ``{SWAG}: A large-scale
  adversarial dataset for grounded commonsense inference,'' in
  \emph{Proceedings of the 2018 Conference on Empirical Methods in Natural
  Language Processing}.\hskip 1em plus 0.5em minus 0.4em\relax Association for
  Computational Linguistics, 2018, pp. 93--104.

\bibitem{Amrita2018DuoRC}
A.~Saha, R.~Aralikatte, M.~M. Khapra, and K.~Sankaranarayanan, ``Duorc: Towards
  complex language understanding with paraphrased reading comprehension,''
  \emph{CoRR}, vol. abs/1804.07927, 2018.

\bibitem{Zhang2018ReCoRD}
S.~Zhang, X.~Liu, J.~Liu, J.~Gao, K.~Duh, and B.~V. Durme, ``Record: Bridging
  the gap between human and machine commonsense reading comprehension,'' 2018.

\bibitem{Talmor2018CommonsenseQA}
A.~Talmor, J.~Herzig, N.~Lourie, and J.~Berant, ``Commonsenseqa: {A} question
  answering challenge targeting commonsense knowledge,'' \emph{CoRR}, vol.
  abs/1811.00937, 2018.

\bibitem{Chen2019CODAH}
M.~Chen, M.~D{'}Arcy, A.~Liu, J.~Fernandez, and D.~Downey, ``{CODAH}: An
  adversarially-authored question answering dataset for common sense,'' in
  \emph{Proceedings of the 3rd Workshop on Evaluating Vector Space
  Representations for {NLP}}.\hskip 1em plus 0.5em minus 0.4em\relax
  Association for Computational Linguistics, 2019, pp. 63--69.

\bibitem{Kwiatkowski2019Natural}
T.~Kwiatkowski, J.~Palomaki, O.~Redfield, M.~Collins, A.~Parikh, C.~Alberti,
  D.~Epstein, I.~Polosukhin, M.~Kelcey, J.~Devlin, K.~Lee, K.~N. Toutanova,
  L.~Jones, M.-W. Chang, A.~Dai, J.~Uszkoreit, Q.~Le, and S.~Petrov, ``Natural
  questions: a benchmark for question answering research,'' \emph{Transactions
  of the Association of Computational Linguistics}, 2019.

\bibitem{Huang2019CosmosQA}
L.~Huang, R.~Le~Bras, C.~Bhagavatula, and Y.~Choi, ``Cosmos {QA}: Machine
  reading comprehension with contextual commonsense reasoning,'' in
  \emph{Proceedings of the 2019 Conference on Empirical Methods in Natural
  Language Processing and the 9th International Joint Conference on Natural
  Language Processing (EMNLP-IJCNLP)}, 2019, pp. 2391--2401.

\bibitem{Clark2019Boolq}
C.~Clark, K.~Lee, M.-W. Chang, T.~Kwiatkowski, M.~Collins, and K.~Toutanova,
  ``{B}ool{Q}: Exploring the surprising difficulty of natural yes/no
  questions,'' in \emph{Proceedings of the 2019 Conference of the North
  {A}merican Chapter of the Association for Computational Linguistics: Human
  Language Technologies, Volume 1 (Long and Short Papers)}.\hskip 1em plus
  0.5em minus 0.4em\relax Association for Computational Linguistics, 2019, pp.
  2924--2936.

\bibitem{Fan2019Eli5}
A.~Fan, Y.~Jernite, E.~Perez, D.~Grangier, J.~Weston, and M.~Auli, ``{ELI}5:
  Long form question answering,'' in \emph{Proceedings of the 57th Annual
  Meeting of the Association for Computational Linguistics}.\hskip 1em plus
  0.5em minus 0.4em\relax Association for Computational Linguistics, 2019, pp.
  3558--3567.

\bibitem{Xiong2019Tweetqa}
W.~Xiong, J.~Wu, H.~Wang, V.~Kulkarni, M.~Yu, S.~Chang, X.~Guo, and W.~Y. Wang,
  ``{TWEETQA}: A social media focused question answering dataset,'' in
  \emph{Proceedings of the 57th Annual Meeting of the Association for
  Computational Linguistics}.\hskip 1em plus 0.5em minus 0.4em\relax
  Association for Computational Linguistics, 2019, pp. 5020--5031.

\bibitem{Liu2019XQA}
J.~Liu, Y.~Lin, Z.~Liu, and M.~Sun, ``{XQA}: A cross-lingual open-domain
  question answering dataset,'' in \emph{Proceedings of the 57th Annual Meeting
  of the Association for Computational Linguistics}.\hskip 1em plus 0.5em minus
  0.4em\relax Association for Computational Linguistics, 2019, pp. 2358--2368.

\bibitem{Gao2019Neural}
J.~Gao, M.~Galley, and L.~Li, ``Neural approaches to conversational ai,''
  \emph{Foundations and Trends® in Information Retrieval}, vol.~13, no. 2-3,
  pp. 127--298, 2019.

\bibitem{Lei2020Conversational}
W.~Lei, X.~He, M.~de~Rijke, and T.-S. Chua, ``Conversational recommendation:
  Formulation, methods, and evaluation,'' in \emph{Proceedings of the 43rd
  International ACM SIGIR Conference on Research and Development in Information
  Retrieval}, ser. SIGIR '20.\hskip 1em plus 0.5em minus 0.4em\relax
  Association for Computing Machinery, 2020, p. 2425–2428.

\bibitem{Zhu2019Learning}
H.~Zhu, L.~Dong, F.~Wei, W.~Wang, B.~Qin, and T.~Liu, ``Learning to ask
  unanswerable questions for machine reading comprehension,'' in
  \emph{Proceedings of the 57th Annual Meeting of the Association for
  Computational Linguistics}.\hskip 1em plus 0.5em minus 0.4em\relax
  Association for Computational Linguistics, 2019, pp. 4238--4248.

\bibitem{Hu2018ReadV}
M.~Hu, F.~Wei, Y.~xing Peng, Z.~X. Huang, N.~Yang, and M.~Zhou, ``Read +
  verify: Machine reading comprehension with unanswerable questions,''
  \emph{ArXiv}, vol. abs/1808.05759, 2018.

\bibitem{Aliannejadi2019Asking}
M.~Aliannejadi, H.~Zamani, F.~Crestani, and W.~B. Croft, ``Asking clarifying
  questions in open-domain information-seeking conversations,'' in
  \emph{Proceedings of the 42nd International ACM SIGIR Conference on Research
  and Development in Information Retrieval}.\hskip 1em plus 0.5em minus
  0.4em\relax Association for Computing Machinery, 2019, p. 475–484.

\bibitem{Du2017Learning}
X.~Du, J.~Shao, and C.~Cardie, ``Learning to ask: Neural question generation
  for reading comprehension,'' in \emph{Proceedings of the 55th Annual Meeting
  of the Association for Computational Linguistics (Volume 1: Long
  Papers)}.\hskip 1em plus 0.5em minus 0.4em\relax Vancouver, Canada:
  Association for Computational Linguistics, 2017, pp. 1342--1352.

\bibitem{Duan2017Question}
\BIBentryALTinterwordspacing
N.~Duan, D.~Tang, P.~Chen, and M.~Zhou, ``Question generation for question
  answering,'' in \emph{Proceedings of the 2017 Conference on Empirical Methods
  in Natural Language Processing}.\hskip 1em plus 0.5em minus 0.4em\relax
  Copenhagen, Denmark: Association for Computational Linguistics, 2017, pp.
  866--874. [Online]. Available:
  \url{https://www.aclweb.org/anthology/D17-1090}
\BIBentrySTDinterwordspacing

\bibitem{Zhou2017Neural}
Q.~Zhou, N.~Yang, F.~Wei, C.~Tan, H.~Bao, and M.~Zhou, ``Neural question
  generation from text: {A} preliminary study,'' \emph{CoRR}, vol.
  abs/1704.01792, 2017.

\bibitem{Pan2019Recent}
L.~Pan, W.~Lei, T.~Chua, and M.~Kan, ``Recent advances in neural question
  generation,'' \emph{CoRR}, vol. abs/1905.08949, 2019.

\bibitem{Chen2020Open}
C.~C. Chen~Qu, Liu~Yang \emph{et~al.}, ``Open-retrieval conversational question
  answering,'' \emph{CoRR}, vol. abs/2005.11364, 2020.

\end{thebibliography}
